\documentclass[11pt]{article}
\usepackage[colorlinks=true,pdfborder=001,linkcolor=blue,citecolor=blue]{hyperref}
\usepackage{url}
\usepackage{stackengine}
\usepackage{nccmath}
\usepackage{mathtools}

\usepackage{mathrsfs}
\usepackage[title]{appendix}
\usepackage{adjustbox}
\usepackage{subcaption}
\usepackage{pgfplots}
\usepackage{tikz}
\usepackage{tikz,fullpage}
\usepackage{rotating}
\usepackage{tabu}
\usepackage{hhline}
\usepackage{multirow}
\usepackage{float}
\restylefloat{table}
\usepackage{adjustbox}
\usepackage[round]{natbib}
\usepackage{booktabs}
\usepackage{array}
\usepackage{tabularx}
\usepackage[utf8]{inputenc}
\usepackage{amssymb,amsmath}
\usepackage{amsfonts}
\usepackage{latexsym}
\usepackage[english]{babel}
\usepackage[T1]{fontenc}
\usepackage{eepic}
\usepackage{epic}
\usepackage{epsfig}
\usepackage{graphics,color}
\usepackage{verbatim}
\usepackage{lscape}
\usepackage{amsthm}
\usepackage{amsmath}
\usepackage{placeins}
\usepackage{float}
\usepackage{caption}
\usepackage{color}
\usepackage{setspace}
\usepackage[ruled,linesnumbered]{algorithm2e}
\usepackage{frcursive}

\usepackage{mathrsfs}
\usepackage{soul}
\usepackage{tikz}
\usepackage{tikz,fullpage}
\usetikzlibrary{arrows,petri,topaths,automata}
\usepackage[short,nocomma]{optidef}
\usepackage{tcolorbox}
\usepackage{fancyvrb}
\tcbuselibrary{skins,breakable}
\newtcolorbox{promptbox}[1][]{%
  enhanced, breakable,
  colback=gray!5, colframe=gray!50!black,
  boxrule=0.5pt, arc=2pt,
  left=6pt, right=6pt, top=4pt, bottom=4pt,
  fonttitle=\bfseries\small,
  fontupper=\small,
  #1%
}

\usepackage[colorinlistoftodos]{todonotes}

\makeatletter

\newif\ifbold
\boldfalse
\boldtrue
\newcommand{\bbf}{\ifbold\bgroup\bf\fi}
\newcommand{\ebf}{\ifbold\egroup\fi}

\renewcommand{\textbf}[1]{\begingroup\bfseries\mathversion{bold}#1\endgroup}
\makeatletter    
\renewcommand{\section}{\@startsection {section}{1}{\z@}%
             {-2ex \@plus -1ex \@minus -.2ex}%
             {1ex \@plus.2ex}%
             {\normalfont\Large\rmfamily\bfseries}}
\renewcommand{\subsection}{\@startsection{subsection}{2}{\z@}%
             {-1.25ex\@plus -1ex \@minus -.2ex}%
             {.75ex \@plus .2ex}%
             {\normalfont\large\rmfamily\bfseries}}
\def\@listI{\leftmargin\leftmargini       
            \parsep .25ex \@plus .1ex     
            \topsep .25ex \@plus .1ex     
            \itemsep \parsep}
\let\@listi\@listI
\@listi
\makeatother
\makeatletter

\@addtoreset{equation}{section}
\makeatother
\makeatletter

\@addtoreset{table}{section}
\makeatother

\usepackage[margin=1in]{geometry}
\usepackage{graphicx}
\graphicspath{ {./Figures/} }
\definecolor{purple}{rgb}{0.4,0.2,1}

\usepackage{titling}

\title{
\LARGE\bf Democratizing Large-Scale Re-Optimization with LLM-Guided Model Patches
\vspace{1ex}
}

\author{\large Tinghan Ye,$^{1,*}$ Arnaud Deza,$^{1}$ Ved Mohan,$^{1}$ El Mehdi Er Raqabi,$^{1,2}$ Pascal Van Hentenryck$^{1}$\\
\footnotesize$^1$\emph{H. Milton Stewart School of Industrial and Systems Engineering, Georgia Institute of Technology, Atlanta, USA}\\
\footnotesize$^2$\emph{Department of Operations and Decision Systems, Université Laval, Québec, Canada}\\
\footnotesize$^*$\emph{Corresponding Author: joe.ye@gatech.edu}\\
}    

\date{}

\begin{document}
\maketitle

\vspace{2cm}
\begin{abstract}
\vspace{0.5cm}

Optimization models developed by operations research (OR) experts are often deployed as decision-support systems in industrial settings. However, real-world environments are dynamic, with evolving business rules, previously overlooked constraints, and unforeseen perturbations. In such contexts, end users (e.g., operators or planners) should ideally re-optimize models to recover feasible and implementable solutions, often without access to the original model developers. In general, however, they often work with expert modelers to apply necessary changes, inducing delays and preventing real-time applications.

This paper introduces an agentic re-optimization framework in which a large language model (LLM) acts as an OR expert, dynamically supporting end users through natural-language interaction. The LLM translates user prompts into structured updates of the underlying optimization model, selects suitable re-optimization techniques from an optimization toolbox, and solves the resulting instance to return implementable solutions. The toolbox leverages primal information, including historical solutions, valid inequalities, solver configurations, and metaheuristics, to accelerate re-optimization while preserving solution quality.  The proposed framework enables interactive and continuous adaptation of deployed optimization models, reducing dependence on OR experts, and improving the sustainability of decision-support systems. 

Extensive experiments on two complementary large-scale real-world case studies demonstrate the effectiveness and scalability of the proposed framework. The first considers online supply chain re-optimization, where solutions must be generated rapidly while remaining close to the deployed plan, whereas the second focuses on offline university exam scheduling, where solution quality is prioritized over runtime. Together, these settings highlight the ability of the framework to support both real-time operational recovery and high-quality planning. Results show that the toolbox-driven architecture significantly improves computational efficiency through primal-based and solver-aware re-optimization techniques, while the structured patch-based updates improve interpretability and traceability of model modifications.
 
\vspace{0.2cm}
\par
\vspace{0.2cm}

\end{abstract}    

\setlength{\parindent}{1em}
\setlength{\parskip}{0.5em}
\doublespacing
\newpage

\section{Introduction}\label{section:1}

Modern business environments are increasingly complex and dynamic, challenging the traditional lifecycle of optimization models in operations and management \citep{erraqabi5}. Classical operations research (OR) practice relies on expert modelers to translate real-world problems into mathematical formulations, calibrate them to data, and deliver executable tools to organizations. Once deployed, however, these models often lose relevance as operational requirements and data landscapes evolve. Each adaptation typically requires renewed expert intervention, an unsustainable process that limits the long-term impact of even the most sophisticated optimization models. In practice, even small operational adjustments can require nontrivial reformulation efforts, making continuous human maintenance both costly and error-prone in large-scale systems.

Given the dynamism and complexity, businesses frequently face rapidly changing business rules, evolving requirements, and unexpected perturbations. This impacts the developed OR models in various ways, including localized updates to constraints, variables, and model parameters required to maintain operational feasibility and decision quality. As a result, users must re-optimize models continuously to maintain solution quality and operational feasibility. Current tools, however, rarely equip users to adjust models autonomously. This creates an important organizational bottleneck, where the ability to adapt deployed optimization systems becomes limited by the availability of specialized modeling expertise and the latency associated with repeated model maintenance cycles. Importantly, this re-optimization task is not merely a re-solving exercise, but a structured model-editing problem that requires understanding how local changes propagate through a tightly coupled optimization system.


Although OR models are often sophisticated, they are inherently generic and do not account for case-specific adaptations. When deployed, end users encounter new or previously overlooked business rules that require immediate adjustments (see e.g., \cite{erraqabi5}). The conventional workflow, returning to the modeler for each update, creates delays and reduces practical value. Although organizations can manually implement common update pathways, maintaining exhaustive adaptation logic becomes increasingly difficult as optimization systems scale, evolve, and interact with diverse operational scenarios. Addressing this gap calls for frameworks that enable users to modify and dynamically re-optimize models in line with evolving operational rules. This is nowadays possible with the recent advances in large language models (LLMs). The LLMs offer new opportunities for interactive and natural-language-based problem-solving. While LLMs have been applied mainly to code generation or decision support, their potential to sustain and evolve formal re-optimization models in large-scale real-world settings remains largely unexplored \citep{li2023large,kong2025alphaopt,simchi2025democratizing}. However, a key limitation of existing approaches is that they typically treat re-optimization as either (i) pure code generation or (ii) black-box repair, without explicitly reasoning over the underlying mathematical structure of the model, the feasibility implications of edits, or the validation and traceability requirements necessary for deployment in operational environments, particularly in \emph{large-scale} settings where even minor modifications can have significant computational and structural impact. This gap is fundamental since OR experts are still required because understanding \emph{what to change} in a large-scale mixed-integer program (MIP) is often more difficult than solving it, exposing a key \emph{interpretability} bottleneck. Even experienced OR practitioners must trace constraint interactions, variable dependencies, and feasibility boundaries under tight time constraints, a task that becomes increasingly impractical at an industrial scale due to model size and structural coupling.


This paper proposes an interactive agentic framework, \textbf{ReOpt-LLM}, in which an LLM bridges the OR model and the end user, serving as an interactive orchestration layer for model adaptation and re-optimization. Once the core model is provided by the OR expert, the LLM manages case-specific adaptations by translating user inputs into model updates and coordinating re-optimization. This setup allows users to interact dynamically with the model, updating parameters, variables, and constraints in real time without necessarily having OR expertise. This is inspired by and aligns with practice, where many business rules are highly customized or may not have been considered during model development by the OR expert. Consequently, the delivered OR tool must allow continuous operation and adaptation to emerging updates without constant expert intervention. \textbf{ReOpt-LLM} leverages the complementary strengths of OR experts and LLMs. Each prompt is translated into changes in model parameters, variables, or constraints, and the LLM serves as an interactive intermediary between the user and the OR model, enabling dynamic re-optimization that recovers and maintains decision quality. One of the main strengths of the proposed framework is its interaction with a re-optimization \emph{toolbox}, which contains primal information and various OR techniques, including historical solutions, solver configurations, valid inequalities, and domain-specific heuristics. This toolbox supports quick re-optimization following a prompt by the user. Importantly, each modification remains explicit, traceable, and verifiable through structured model patches and validation steps, allowing users to inspect how requests are translated into optimization updates and how the resulting solutions satisfy the intended operational changes. A central design principle is that the LLM does not replace optimization expertise, but instead operationalizes it by orchestrating a structured set of solver-aware actions, thereby reducing the need for direct human modeling intervention while preserving mathematical rigor.


Throughout \textbf{ReOpt-LLM}, the aim is to advance the sustainability of decision-support systems in real-world settings, reducing dependence on continuous expert intervention while ensuring long-term alignment with changing operational realities. This also contributes to the emerging literature on LLM-assisted decision making, demonstrating how language-based intelligence can bridge the gap between model formulation and practical implementation. By embedding re-optimization capabilities directly within an LLM-guided orchestration framework, the proposed approach transforms deployed optimization models from static analytical artifacts into continuously adaptive decision-support systems. A key additional insight is that the effectiveness of such systems is not uniform across problem scales. While simpler or well-conditioned instances can often be handled with minimal intervention, large-scale structured MIPs require careful coordination among model editing, heuristic guidance, and solver configuration, motivating the proposed toolbox-driven design.


The contributions of this paper are the following:
\begin{enumerate}
    \item \textbf{Agentic Re-optimization Framework.} A dynamic and interactive re-optimization framework connecting the OR model, the re-optimization toolbox, and the end user, enabling continuous integration of user input and adaptation to diverse operational contexts.
    \item \textbf{OR Perspective.} LLM-assisted re-optimization by translating user queries into mathematical model updates. Once the mathematical model is updated, the LLM selects \emph{suitable} re-optimization modules from the toolbox and solves the resulting instance using these modules and an optimization solver. The toolbox is a key driver of scalability, enabling efficient large-scale re-optimization through reusable primal information, warm starts, solver configurations, valid inequalities, and customized heuristics. This design allows the framework to handle large-scale instances under strict time budgets while maintaining solution quality.
    \item \textbf{LLM Perspective.} Methods to prompt, guide, and structure the LLM for reliable re-optimization, including structured patch-based model edits, toolbox-aware decision making, and failure analysis under model-edit ambiguity. These patch operations improve interpretability and deployment reliability by making every modification to the optimization model explicit, traceable, verifiable, and reviewable by the end user, rather than implicit or black-box. 
    \item \textbf{Applications.} Extensive experiments on two real-world large-scale problems highlight the ability of the framework to re-optimize reliably and return satisfactory solutions within reasonable execution time. The two case studies are complementary. The OCP Group case represents an \emph{online re-optimization} setting, where perturbations occur during the execution of a supply chain, decisions must be updated under strict time constraints, and new solutions must remain as close as possible to the already deployed plan, making both computational speed and solution stability critical. In contrast, the Cornell exam scheduling case corresponds to an \emph{offline re-optimization} setting, where the schedule is iteratively refined with the university registrar before publication, and feasibility and solution quality are prioritized over runtime. Together, these two settings demonstrate the robustness of the proposed framework across both real-time operational decision-making and pre-deployment planning environments.
\end{enumerate}

The remainder of the paper is organized as follows. Section~\ref{section:2} reviews relevant literature. Section~\ref{section:3} states the problem setting. Section~\ref{section:4} presents the \textbf{ReOpt-LLM} framework. Section~\ref{section:5} provides the mathematical formalization and the implementation. Section~\ref{section:6} illustrates a toy example. Section~\ref{section:eval-protocol} describes the shared evaluation protocol used in the case studies. Sections~\ref{section:ocp} and~\ref{section:cornell} present two large-scale real-world applications. Section~\ref{section:conclusions} concludes with a discussion of broader implications, challenges, and directions for future research.

\section{Literature Review}\label{section:2}

This section reviews prior work relevant to this research. It first examines the emerging role of LLMs in optimization, then discusses re-optimization techniques in OR, and finally positions this work relative to these streams.

\subsection{LLMs for Optimization}

The application of LLMs to the field of optimization represents a nascent yet rapidly advancing area of research, spanning model formulation, solution support, human-in-the-loop decision making, and optimization education. Several recent position papers, surveys, and case studies have provided foundational perspectives on the emerging role of LLMs in various optimization contexts and their potential to democratize solution modeling \citep{li2023large, simchi2025democratizing, simchi2025large, chen2025optimind, xiao2025survey}. Research in this area explores several distinct roles for LLMs across the optimization process. This distinction is useful for positioning this work, which focuses on the post-deployment stage where an already validated optimization model must be continuously adapted and re-optimized under evolving operational conditions.

A primary application, NL4OPT (natural language for optimization), focuses on the modeling and formulation phase. This involves converting unstructured natural language descriptions into formal optimization formulations~\citep{ramamonjison2023nl4opt}. This task is typically approached using prompt-based techniques~\citep{xiao2023chain, ahmaditeshnizi2024optimus, kong2025alphaopt, liang2025llm, zhang2025hierarchical} or learning-based methods~\citep{jiang2024llmopt, huang2025orlm, zhou2025auto}. Related work also investigates using LLMs for other pre-solving tasks, such as configuring solver algorithms~\citep{lawless2025llms}, and post-processing tasks, like filtering a pool of Pareto-optimal solutions through a multi-objective lens~\citep{jovine2025listen}. Closely related, OptiChat~\citep{chen2025optichat} provides a conversational interface to an already-built optimization model for interpretation, infeasibility diagnosis, sensitivity analysis, and transient what-if evaluation. \citet{drossman2026let} study a related conversational setting in which an agent makes bounded edits to an already-deployed formulation in response to stakeholder preferences, such as fixing a variable value or imposing an objective bound. In that setting, the underlying decision space remains unchanged across the conversation. These works primarily address the construction of optimization models from initial problem descriptions, the explanation of an existing one, or the elicitation of stakeholder preferences over one. This paper also starts from a validated deployed model, but it targets operational changes that arise after deployment. Here, user requests introduce new operational adaptations or previously uncaptured requirements that may change the feasible region. The agent translates those requests into structural updates of the model's parameters, variables, constraints, or objective, and then applies solver-aware re-optimization techniques instead of solving the revised model from scratch.

Beyond problem formulation, another research thrust integrates LLMs directly into the solution process. This ranges from positioning the LLM as a direct, formulation-free optimizer~\citep{yang2023large} to embedding it as an active component within traditional algorithms. Spurred by advances in LLM-based code generation \citep{novikov2025alphaevolve}, these more integrated approaches use LLMs to dynamically generate or select mathematical heuristics \citep{romera2024mathematical}, to act as sophisticated crossover or mutation operators in evolutionary algorithms \citep{ye2024reevo}, or to guide the selection policy in Monte Carlo Tree Search \citep{zheng2025monte}. ReOpt-LLM differs from this stream in that the LLM is not used as the optimizer itself or as an unconstrained heuristic generator; rather, it acts as an OR-aware orchestration layer that coordinates structured model adaptation, validation, and solver-aware re-optimization techniques.

LLMs are also emerging as a novel mechanism for managing the human-in-the-loop aspect of optimization, particularly for preference elicitation. Their capacity for natural language interpretation enables zero-shot preference modeling, allowing a decision-maker's objectives to be inferred directly from verbal descriptions rather than through explicit pairwise queries. To date, this research has primarily employed LLMs as auxiliary components to guide interactive questioning \citep{Lawless2023IWantItThatWay, Austin2024PEBOL}, extract latent preferences from unstructured text \citep{Bang2025PURE}, or serve as simulators of user behavior \citep{Okeukwu2025Community, Zhang2025UR4Rec}. These studies use language primarily to elicit, infer, or simulate preferences over decisions, whereas the focus in this paper is on operational updates that alter the feasible region, objective, or model data and therefore require re-solving the modified optimization problem.

Finally, complementing these application-oriented efforts, a separate stream of work evaluates the foundational knowledge of LLMs on core optimization principles. For instance, studies have assessed their understanding of primal-dual theory, which is crucial for ensuring reliable modeling and for their potential use in education~\citep{klamkin2025dualschool}.

\subsection{Re-optimization Techniques}

Re-optimization plays a central role in maintaining robustness and adaptability in environments subject to continuous or unexpected change. It is commonly divided into two broad categories. \emph{Major re-optimization} refers to infrequent but substantial redesigns triggered by significant disruptions. These are typically strategic or tactical in nature and are often resolved from scratch using exact optimization methods. In contrast, \emph{minor re-optimization} deals with routine or operational disturbances that occur frequently, sometimes in real time. In this setting, solutions are not rebuilt entirely. Instead, previously computed solutions are adapted using heuristics or approximation schemes that exploit existing structure.

From the perspective of large-scale or structural re-optimization, \citet{chen2012resilience} introduce a stochastic mixed-integer formulation to assess system resilience and to determine recovery actions after disruptions. Similarly, \citet{bruno2021reorganizing} study the redesign of a postal collection network in Italy, where the decision problem involves selecting urban postboxes to deactivate. They formulate the problem as a mixed-integer program and solve it using a two-stage exact approach.

Minor re-optimization, on the other hand, focuses on fast adaptation to localized changes. For example, \citet{d2010running} consider adjustments to train running profiles using a graph-based representation combined with constructive heuristics that preserve feasibility while modifying existing schedules. \citet{archetti2013reoptimizing} study variations of the Rural Postman Problem and propose heuristic strategies to update solutions when edges are added or removed. Within a more general combinatorial framework, \citet{schieber2018theory} formalize combinatorial re-optimization problems and develop methods that explicitly control transition costs between solutions.

In operational settings, \citet{dong2018reoptimization} investigate maritime inventory routing under uncertainty using a rolling-horizon scheme solved with a commercial MIP solver. Evolutionary approaches are examined by \citet{doerr2019fast}, who show that standard population-based methods can be inefficient in dynamic environments and propose mechanisms based on diversity preservation to improve adaptability. In workforce scheduling, \citet{hassani2020real} and \citet{hasani2021multi} design fast heuristic procedures that generate near-Pareto solutions by explicitly accounting for cost and schedule deviations, often embedding neighborhood structures within integer programming formulations. More recently, \citet{wu2024towards} propose a general framework for resilience-driven re-optimization, where disruptions and recovery actions are jointly modeled. Their approach leverages primal solutions through fixing strategies, warm starts, valid inequalities, and machine learning components, and is validated on large-scale industrial instances.

Recent work relying on artificial intelligence (AI) highlights the challenge of deciding when to re-solve optimization models in real time. The proximal policy optimization (PPO) framework addresses this by learning optimal re-solving times, balancing solution quality with computational cost in dynamic MIP settings \citep{ai2025solve}.

\subsection{Paper Positioning}

This paper lies at the intersection of OR, AI, and business analytics, addressing the challenge of sustainable model-driven decision-support systems. It proposes a framework that assumes that {\em OR experts developed and validated a mathematical optimization model, which is then deployed to end users}. In practice, the main difficulty is no longer model formulation or optimization, but the continuous and correct adaptation of a large, tightly coupled MIP under evolving operational conditions, where even small edits can have non-local and non-obvious feasibility and optimality implications. This makes re-optimization fundamentally a reasoning problem over model structure rather than a mere computational task. The LLM agent (i) interprets user queries (in natural language) containing operational perturbations, evolving operational requirements, and other structural updates, (ii) translates them into corresponding model modifications, (iii) picks suitable generic and/or domain-specific re-optimization techniques, (iv) re-optimizes the model under new or perturbed conditions by calling an optimization solver enhanced with the re-optimization techniques, and (v) delivers actionable insights in an \emph{intuitive} form, i.e., a form that the end user can understand and implement. A key distinction from prior work is that these steps are not treated as independent tasks, but as a tightly coupled closed-loop decision system in which model editing, solver configuration, and primal enhancement interact. Unlike prior LLM-based optimization approaches that focus primarily on model generation, code synthesis, or black-box solution repair, this framework explicitly operates on structured, auditable, and reviewable model edits supported by explicit validation and traceability mechanisms, and integrates solver-aware re-optimization decisions grounded in OR methodology. This is crucial because naive code-level modifications are insufficient to guarantee feasibility or meaningful control over combinatorial structures in large-scale MIPs. By embedding re-optimization capabilities within the LLM, \textbf{ReOpt-LLM} transforms traditional optimization tools from static artifacts into continuously adaptive and collaborative decision-support systems that evolve through human–AI collaboration. Importantly, the LLM is not positioned as a solver or optimizer replacement, but as an {\em OR-aware orchestration layer} that externalizes expert reasoning (model diagnosis, modification selection, and solver guidance) in a way that is both scalable and interpretable. This allows domain users to interact with complex optimization systems without requiring deep expertise in mathematical modeling, while preserving deployment reliability through solver validation, structured feedback, and explicit human review of model modifications. Overall, the framework views re-optimization as a structured decision-understanding problem, where language, model structure, and primal information are jointly leveraged to sustain feasibility, operational responsiveness, and decision quality under continuous change. To the best of the authors' knowledge, no prior work unifies structured model editing, solver-aware re-optimization, and primal-toolbox selection within a closed-loop LLM-driven framework for large-scale re-optimization.

\section{Problem Description}\label{section:3}

Consider a generic optimization setting in which an OR model is developed, validated, and deployed to support decision-making in a dynamic and complex environment. The deployed model constitutes a well-defined computational object that encodes the operational logic of the system. Let $x$ denote the vector of decision variables, partitioned into continuous and integer components, and consider the following MIP formulation:

\begin{equation}
\label{model}
\min_{x}\left\{ c^\top x \;\middle|\; Ax \le b,\; x \in \mathbb{R}^{n} \times \mathbb{Z}^{m} \right\},
\end{equation}

\noindent where $c$ is the cost vector, $A$ is the constraint matrix, and $b$ is the right-hand side vector. This formulation captures a wide range of decision-support applications, including production planning, scheduling, transportation, and resource allocation. This work assumes that the underlying problem structure remains a minimization problem of the form \eqref{model}, while its components may evolve over time. Consequently, the MIP is not static, but a parameterized and structured model whose elements may be modified as the environment changes.

In practice, such models are often large-scale and highly structured, involving thousands to millions of variables and constraints. They are typically designed and calibrated by OR experts in collaboration with domain specialists, and then deployed as decision-support tools within organizations. Once deployed, these models are used repeatedly to generate tactical or operational plans under varying data inputs. However, real-world environments are inherently dynamic, and the conditions under which the model was originally designed may no longer hold. Over time, perturbations arise, operational requirements evolve, and previously uncaptured constraints or structural adjustments emerge. These changes may affect any component of the model, including the cost vector $c$, the decision variables $x$, the constraint matrix $A$, or the right-hand side vector $b$. For example, new constraints may need to be introduced, existing constraints modified or removed, variable domains restricted or expanded, or parameter values updated. Crucially, such modifications are often described locally in natural language, yet they induce global and highly non-trivial effects on feasibility and optimality. In practice, these updates are often initiated by planners, operators, or other domain users who understand the operational context but lack expertise in mathematical optimization. Consequently, effective re-optimization requires not only computational efficiency but also interactive mechanisms that allow users to express intended operational changes, inspect the resulting model modifications, and validate that the revised solution aligns with operational objectives and constraints.

As a result, the deployed model must be continuously re-optimized to remain relevant and effective. Importantly, this re-optimization process goes beyond simply re-solving the model with updated data. It requires modifying the mathematical structure of the optimization problem itself. Ideally, these modifications should remain explicit and inspectable, allowing users to understand how localized operational requests alter the underlying optimization model. From a computational perspective, this is challenging because even small changes in the model components can significantly alter the feasible region and the solution space, especially for large-scale models. {\em From a practical perspective, this is difficult because end users typically lack the expertise required to manipulate mathematical models directly, and repeated reliance on OR experts is costly and slow.} Although some recurring updates can be manually encoded within decision-support software, exhaustively anticipating and maintaining all possible operational adaptations becomes increasingly difficult as optimization systems scale, evolve, and interact with diverse operational scenarios. Moreover, these updates must be performed quickly to support real-time or near-real-time decision-making, creating a tension among model expressiveness, solution quality, and computational efficiency. Delays in adapting deployed optimization systems can significantly reduce operational responsiveness and diminish the practical value of model-driven decision-support tools. A central difficulty is that identifying what structural edits preserve feasibility while improving or maintaining solution quality is often strictly more difficult than solving the resulting optimization problem itself. Therefore, even in the absence of language models, enabling systematic, reliable, and fast re-optimization of evolving large-scale MIP models is a challenging and important problem.

This setting motivates the need for a framework that can (i) interpret user-specified changes expressed in natural language, (ii) translate these changes into precise, reviewable, and verifiable updates of the mathematical model \eqref{model}, and (iii) efficiently re-optimize the resulting instance while preserving solution quality and implementability. The next section presents an LLM-assisted agentic framework designed to address this challenge.

\section{Agentic Re-optimization Framework}\label{section:4}

\begin{figure}[!t]
    \centering
    \includegraphics[width=\linewidth]{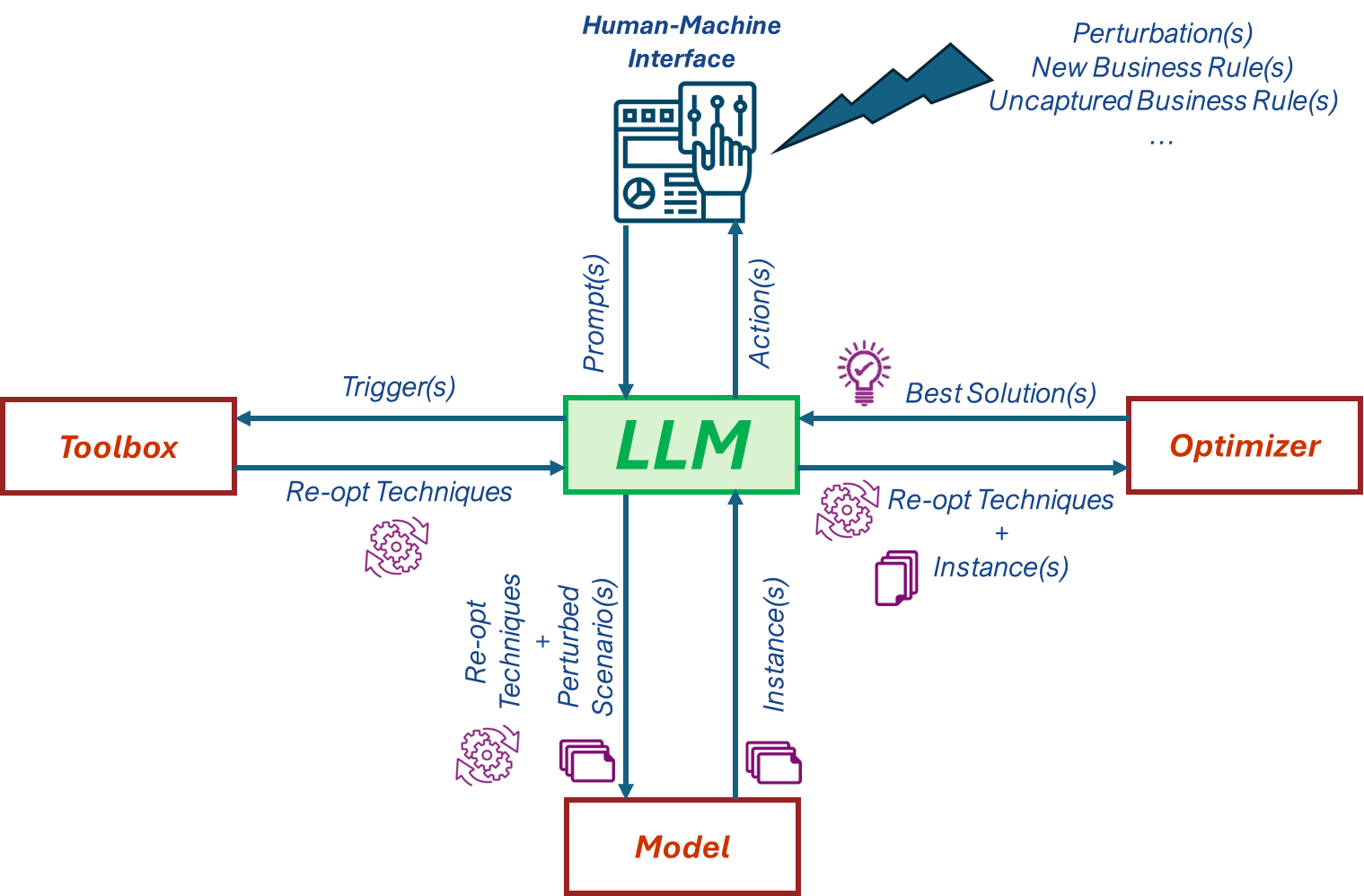}
    \caption{\textbf{ReOpt-LLM} Framework}
    \label{fig:collaborative_framework}
\end{figure}

\textbf{ReOpt-LLM}, the proposed LLM-assisted collaborative re-optimization framework, is illustrated in Figure \ref{fig:collaborative_framework}; it integrates the end user, the optimization model, and the optimizer through a language-based interface. The optimization model is assumed to have been developed and validated by OR expert(s) in collaboration with the end user's organization (\emph{Step 0}). As the business environment evolves, e.g., due to perturbations or changing operational requirements, the existing model may no longer produce valid solutions, necessitating re-optimization (\emph{Step 1}). The user communicates these emerging issues via natural-language prompts to the LLM (\emph{Step 2}), which translates them into corresponding changes in the model parameters and variables (\emph{Step 3}). The updated model is then instantiated to generate new problem instances (\emph{Step 4}), which are solved by an optimizer to produce feasible and high-quality solutions (\emph{Step 5}). The optimizer uses re-optimization techniques from the toolbox to enhance the solver. The re-optimized solution is returned to the user, forming a continuous feedback loop that combines human expertise, AI, and optimization for real-time, adaptive decision support. The rest of the section describes \textbf{ReOpt-LLM} in detail, presenting the step-by-step operation, the LLM trigger mechanism, workflow, and role, the metrics used to evaluate LLM performance, and the speed and optimization techniques.

\subsection{Step-by-step Description}

This section describes the seven steps involved in the \textbf{ReOpt-LLM} framework.

\noindent \textbf{Step 0 – Model Validation and Delivery.} The initial optimization model is developed by OR expert(s) and iteratively refined in collaboration with the end user's organization. Through repeated validation and testing, the model is calibrated to capture the company’s operational logic, business rules, and performance objectives. Once finalized, the model is deployed within the organization for the end user, who can leverage it to optimize operations and, when needed, re-optimize as long as the underlying assumptions remain valid. This step ensures that the model reflects both technical rigor and practical relevance. In the absence of perturbations, the end user uses the model in the normal loop, which starts from the \emph{Human-Machine Interface} and goes to the \emph{Model} through the dashed arrow in Figure \ref{fig:collaborative_framework}.

\noindent \textbf{Step 1 – Identification of Changes.} Over time, the operational environment may change, introducing perturbations, new business rules, and/or previously uncaptured requirements. These changes can occur at any moment throughout the planning horizon, whether in real time, daily, or weekly, reflecting the continuous evolution of operations. {\em ReOpt-LLM is designed to incorporate all such emerging changes dynamically, ensuring continuous model adaptability.} When the existing model can no longer produce feasible or optimal solutions under these conditions, the end user recognizes the need to update and re-optimize, triggering the collaborative re-optimization process.

\noindent \textbf{Step 2 – User Prompting via LLM.} The end user communicates emerging operational challenges to the LLM through a natural-language interface. The prompt may describe new rules, exceptions, or situational changes in the business environment. By allowing the user to express requirements in natural language, the system reduces the barrier to model adaptation and encourages iterative interaction between human expertise and computational reasoning. 

\noindent \textbf{Step 3 – Translation of Prompts into Model Updates.} The LLM interprets the user’s prompts and translates them into precise modifications of the optimization model. Specifically, the LLM identifies the effects on the model components, including the cost vector, decision variables, right-hand side vector, and constraint matrix, while assuming that the underlying problem structure (e.g., a minimization problem) remains unchanged. The LLM thus produces an updated scenario that captures the operational impact of newly identified rules or perturbations. The resulting modifications remain explicit and inspectable, allowing the end user to review how the operational request alters the optimization model before re-optimization is performed.

\noindent \textbf{Step 4 – Toolbox Selection.} Based on the interpreted changes, the LLM selects appropriate generic and domain-specific re-optimization techniques from the toolbox. This step operationalizes OR expertise within the collaborative re-optimization workflow by mapping structural model modifications to suitable primal and algorithmic strategies. Depending on the nature of the perturbation, the LLM may activate warm-start mechanisms, exploit historical primal solutions, introduce valid inequalities, adjust solver parameters, or invoke specialized heuristics. These selections explicitly condition both the instance construction and the downstream optimization process, ensuring that re-optimization is informed by problem structure rather than treated as a black-box solve.

\noindent \textbf{Step 5 – Instance Generation.} Using the modified model, new problem instances are generated that reflect the updated operational environment. These instances formalize the re-optimization task, ensuring that all relevant changes are incorporated into a solvable mathematical representation. This step bridges the translation performed by the LLM and the computational optimization process.

\noindent \textbf{Step 6 – Optimization and Solution Feedback.} The optimizer solves the updated instances, producing feasible and high-quality solutions under the revised conditions. The results are communicated, in the form of actionable decisions, back to the user, who can evaluate and implement them. The framework can additionally provide structured feedback describing the applied model updates, activated re-optimization strategies, and validation outcomes, helping users assess whether the generated solution aligns with the intended operational adaptation. This creates a continuous human-in-the-loop feedback process in which human expertise, AI, and optimization interact dynamically, supporting adaptive and sustainable decision-making in evolving operational contexts.

\subsection{LLM Trigger Mechanism, Workflow, and Role}

The re-optimization process is triggered whenever a user prompt implies a change in the underlying optimization problem. Any such prompt can be mapped to modifications in one or more of the core components that define the model: the cost vector, the decision variables, the constraint matrix, or the right-hand side vector. Notably, updates to these components allow the framework not only to adjust existing constraints but also to add new constraints or remove obsolete ones, enabling flexible adaptation of the model to evolving operational requirements. By framing the problem in this way, every user input becomes structured and actionable, providing a structured basis for the LLM to generate updates and for systematic evaluation of its performance. This structure also supports validation and human review before re-optimization, helping ensure that the generated modifications correctly reflect the intended operational adaptation.

To operationalize the framework, the LLM must first understand the \emph{what}, i.e., the specific emerging issues, business rules, or perturbations described by the user. Once these impacts are identified, it determines how to update the scenario(s), translating the natural-language description into precise modifications of the cost vector, decision variables, right-hand side, and constraints. With the updated scenario defined, the LLM constructs the corresponding updated optimization instance, ensuring that it accurately reflects the new operational conditions. This model is then submitted to the optimizer, which computes feasible solutions. Finally, the resulting solution is returned to the user, completing the cycle and enabling continuous, adaptive re-optimization.

\subsection{Evaluation Metrics for the LLM}

The performance of the LLM within the \textbf{ReOpt-LLM} framework can be evaluated along three complementary dimensions. The first dimension assesses the accuracy in identifying model components, i.e., whether the LLM correctly determines which elements, including the cost vector, decision variables, constraint matrix, or right-hand side, require updating in response to the user’s prompt. The second dimension measures the correctness of the updates, i.e., whether the LLM modifies the identified components appropriately to reflect the intended operational changes. The third dimension evaluates the quality and usability of the resulting solutions, including feasibility, optimality, and consistency of the updated model; it also evaluates the relevance of the solutions for the end user, i.e., whether the prompt is correctly addressed. Together, these metrics provide a comprehensive assessment of the LLM’s effectiveness in supporting real-time, adaptive decision-making.


\subsection{Speed and Optimization Techniques}

A key requirement of the \textbf{ReOpt-LLM} framework is that updates and solution generation should occur rapidly, enabling near real-time decision support even as operational conditions evolve. To achieve this, the toolbox can leverage a variety of OR techniques, including exact solvers, metaheuristics, accelerated exact methods, ML-guided approaches, and proxy models. By flexibly selecting or combining these methods, the framework ensures that high-quality solutions are produced efficiently, balancing computational speed with solution accuracy and implementability, while maintaining the responsiveness required for dynamic, continuous re-optimization in practical operational settings.

\section{Mathematical Formalization and Framework Implementation}\label{section:5}

This section presents the mathematical formalization of the \textbf{ReOpt-LLM} framework and its implementation. Subsection~\ref{sec:5-structured} introduces the structured model representation, together with the metadata used by the agents. Subsection~\ref{sec:5-dynamic} formalizes the dynamic setting and the structured events that summarize natural-language changes. Subsection~\ref{sec:5-patch} defines the patch language used to mutate the structured state. Subsection~\ref{sec:5-workflow} describes the collaborative multi-module workflow, in which a single integrated patch planner jointly interprets the request and proposes candidate edits, a deterministic programmer normalizes them, a strategy selector picks a re-optimization strategy, and a validator solves the updated model under a validation-triggered retry loop. Subsection~\ref{sec:5-closedloop} combines these components into a closed-loop algorithm.

Consider a generic optimization problem of the form
\begin{equation}
    \label{eq:base-problem}
    \min_{x \in \mathcal{X}(\mathcal{M}, p)} \;\; f(x; p),
\end{equation}
where $x$ denotes the vector of decision variables, $\mathcal{M}$ denotes the structured model template, $p$ denotes the collection of model parameters (e.g., demands, processing times, exam sets, capacities, costs), and $\mathcal{X}(\mathcal{M}, p)$ is the feasible region defined by the constraints generated from $\mathcal{M}$ and parameterized by $p$. The feasible region is generated from a set of \emph{decision variable families}, \emph{constraint families}, and \emph{objective components}.

\subsection{Structured Model Representation}\label{sec:5-structured}

The optimization model has a structured representation of the form
\[
    \mathcal{M} = (\mathcal{V}, \mathcal{C}, \mathcal{O}),
\]
where $\mathcal{V}$ is a collection of variable families, $\mathcal{C}$ is a collection of constraint families, and $\mathcal{O}$ is a set of named objective components. The parameter vector $p$ is shared across these collections, and $\mathcal{P}$ denotes the set of named parameter entries in $p$ (e.g., \texttt{supply}, \texttt{demand}, \texttt{costs}).

\subsubsection{Decision variable families}
Each variable family $v \in \mathcal{V}$ is associated with a finite index set $\mathcal{I}_v$ and a collection of decision variables
\[
    \{ x_i \}_{i \in \mathcal{I}_v},
\]
where all variables in the family share the same type and interpretation (e.g., flows on arcs, assignment variables, start times). Let $\mathrm{type}(v) \in \{\text{binary}, \text{integer}, \text{continuous}\}$ denote the variable type for family $v$, and let $\ell_i$ and $u_i$ denote the lower and upper bounds for each $x_i$, which may depend on $p$.

\subsubsection{Constraint families}
Each constraint family $c \in \mathcal{C}$ is represented by: (1) \textbf{a finite index set} $\mathcal{I}_c$, (2) \textbf{a left-hand side specification} $g_c(x; p, i)$, and (3) \textbf{a right-hand side specification} $h_c(p, i)$. The instantiated constraints take the form
\[
    g_c(x; p, i) \;\le\; h_c(p, i) \qquad \forall i \in \mathcal{I}_c,
\]
where the sense $\le$  can be replaced by $\ge$ or $=$. Different families correspond to different structural roles (e.g., capacity, balance, precedence, fairness).

\subsubsection{Objective components}
The objective function is decomposed into a finite collection of components indexed by $\mathcal{O}$:
\[
    f(x; p) \;=\; \sum_{k \in \mathcal{O}} w_k \, f_k(x; p),
\]
where $f_k$ encodes a contribution such as transportation cost, tardiness, overtime, or fairness, and $w_k \in \mathbb{R}$ is a tunable weight.

\subsubsection{Metadata for retrieval}
Each retrievable object $\omega \in \mathcal{V} \cup \mathcal{C} \cup \mathcal{O} \cup \mathcal{P}$ carries (1) a short textual description $\mathrm{desc}(\omega)$ and (2) a finite tag set $\mathrm{tags}(\omega)$ (e.g., \texttt{capacity}, \texttt{routing}, \texttt{student}, \texttt{time-window}). These metadata fields do not affect the mathematical model but are exposed to the patch planner to help identify the components and parameter entries relevant to a given change request.

\subsection{Dynamic Setting and Structured Events}\label{sec:5-dynamic}

Over time, new information arrives in two forms that describe parameter or structural modifications in $\mathcal{P}$, $\mathcal{V}$, $\mathcal{C}$, or $\mathcal{O}$:
\begin{itemize}
    \item \textbf{Parameter updates}: the parameter vector $p$ is updated following a change (e.g., revised demands, machine calendars, updated travel times).
    \item \textbf{Natural-language rule updates}: a user or decision-maker expresses a change such as, \emph{``Arc $(i,j)$ is unavailable today''}, \emph{``Machine $m$ is down from 10:00--14:00''}, \emph{``First-year students should not have evening exams''}, \emph{``Product $q$ must be completed by tomorrow''}.
\end{itemize}

Let $\Delta_t$ denote the natural-language description of changes at time $t$. The mutable state is denoted by
\[
    \mathcal{Z}_t = (\mathcal{M}_t, \mathcal{P}_t, p_t),
\]
where $\mathcal{M}_t$ collects variable, constraint, and objective-component families, $\mathcal{P}_t$ denotes the named parameter entries available at time $t$, and $p_t$ collects their current values. For compactness, define the state-dependent feasible region and objective as
\[
    \mathcal{X}(\mathcal{Z}_t) := \mathcal{X}(\mathcal{M}_t, p_t), \qquad F(x;\mathcal{Z}_t) := f(x; p_t).
\]
The goal is to incrementally update $\mathcal{Z}_{t-1}$ to $\mathcal{Z}_t$ to incorporate the change $\Delta_t$, then solve
\[
    x_t^\star \in \arg\min_{x \in \mathcal{X}(\mathcal{Z}_t)} F(x;\mathcal{Z}_t),
\]
efficiently and reliably, while preserving interpretability, correctness, and implementability.
The framework first represents $\Delta_t$ as a structured event
\[
    E_t = \big( S_t, \iota_t, \sigma_t \big),
\]
where $S_t$ is a description of the \emph{affected entities}, encoded as subsets of index sets in the model (e.g., $S_t$ may refer to a subset of machines, arcs, students, or time slots, corresponding to subsets of some $\mathcal{I}_v$ or $\mathcal{I}_c$); $\iota_t \in \{\text{tighten}, \text{relax}, \text{forbid}, \text{prioritize}, \text{update}\}$ is an optional \emph{semantic intention} retained as auxiliary metadata; and $\sigma_t$ is a short textual summary of the requested edit. The structured event $E_t$ is an abstract representation of the change that the integrated planner conditions on $S_t$ and $\sigma_t$ when proposing model edits.

\subsection{Patch Language}\label{sec:5-patch}

All updates to $\mathcal{Z}_t$ are expressed through a restricted \emph{patch language}, i.e., a model-edit domain-specific language (DSL). Each patch is a tuple
\[
    \pi = (\texttt{op}, \texttt{target}, \texttt{scope}, \texttt{update}),
\]
where \texttt{op} is an operation type, \texttt{target} refers to named parameter entries in $\mathcal{P}_t$ and their values in $p_t$ or to one or more elements of $\mathcal{V}_t$, $\mathcal{C}_t$, or $\mathcal{O}_t$, \texttt{scope} identifies relevant indices (e.g., particular machines, arcs, students, time slots), and \texttt{update} specifies the mathematical modification (e.g., new parameter values, new bounds, modified right-hand side, updated weight). After applying a patch $\pi$, the mutable state transitions to
\[
    \mathcal{Z}_t \;=\; \pi(\mathcal{Z}_{t-1}).
\]
Multiple patches generated for the same change request can be grouped into a \emph{candidate action set} that is applied as a unit.

The supported patch-operation vocabulary is summarized in Table~\ref{tab:patch_ops}. Operations on individual indices cover parameters, variable bounds, constraint right-hand sides, left-hand-side specifications, objective coefficients, and objective weights. Pattern-based operations support batch edits over many rows or variables selected by a structural pattern in LP-backed runtimes, which is useful when a single natural-language rule modifies a large family at once. Structural operations create or remove entire families.

The vocabulary is intentionally small. Most operational prompts are handled as data edits, such as changing a demand value, tightening a route bound, or adjusting an objective coefficient; these operations keep the algebraic structure fixed and only mutate existing entries. When a prompt states a relative change, such as ``increase demand by 10,'' the corresponding patch may carry a \texttt{delta} rather than forcing the planner to compute the new absolute value. Pattern-based LP operations are reserved for cases in which the runtime exposes named rows or variables but not a compact structured family. Structural operations are used only when the requested change adds or removes an entire modeling object, such as a new policy constraint family.

\begin{table}[!ht]
\centering
\small
\caption{Supported patch-operation vocabulary. Pattern-based operations apply to a regex-selected subset of LP rows or variables. Structural operations modify $\mathcal{V}$, $\mathcal{C}$, or $\mathcal{O}$.}
\label{tab:patch_ops}
\begin{tabular}{lp{10cm}}
\toprule
\textbf{Operation} & \textbf{Effect} \\
\midrule
\texttt{UPDATE\_PARAMETER} & Replace or additively update a named parameter entry, optionally at a keyed index. \\
\texttt{UPDATE\_BOUND} & Set the lower or upper bound of an existing variable-family member. \\
\texttt{UPDATE\_CONSTRAINT\_RHS} & Replace or additively update the right-hand side of an existing constraint-family row. \\
\texttt{UPDATE\_CONSTRAINT\_LHS} & Replace the left-hand-side specification $g_c(x; p_t, i)$ for an existing constraint family. \\
\texttt{UPDATE\_OBJECTIVE\_COEFF} & Replace or additively update an indexed coefficient inside objective component $f_k$. \\
\texttt{UPDATE\_OBJECTIVE\_WEIGHT} & Replace or additively update the weight $w_k$ of a named objective component. \\
\texttt{UPDATE\_COEFFICIENT} & Modify nonzero LP matrix coefficients selected by variable and constraint-name patterns. \\
\texttt{FIX\_VARIABLES\_BY\_PATTERN} & Set lower and upper bounds for LP variables selected by a name pattern and optional filters. \\
\texttt{UPDATE\_CONSTRAINT\_RHS\_BY\_PATTERN} & Set or scale right-hand sides for LP rows selected by a constraint-name pattern. \\
\texttt{ADD\_VARIABLE\_FAMILY} & Add a new decision-variable family, $\mathcal{V}_t \leftarrow \mathcal{V}_t \cup \{v'\}$. \\
\texttt{ADD\_CONSTRAINT\_FAMILY} & Add a new constraint family, $\mathcal{C}_t \leftarrow \mathcal{C}_t \cup \{c'\}$. \\
\texttt{REMOVE\_CONSTRAINT\_FAMILY} & Remove an existing constraint family, $\mathcal{C}_t \leftarrow \mathcal{C}_t \setminus \{c'\}$. \\
\texttt{ADD\_OBJECTIVE\_COMPONENT} & Add a new objective component, $\mathcal{O}_t \leftarrow \mathcal{O}_t \cup \{k'\}$. \\
\bottomrule
\end{tabular}
\end{table}

\subsection{Agentic Workflow and Retry Loop}\label{sec:5-workflow}

\begin{figure}[!ht]
      \centering                                                        
      \includegraphics[width=\linewidth]{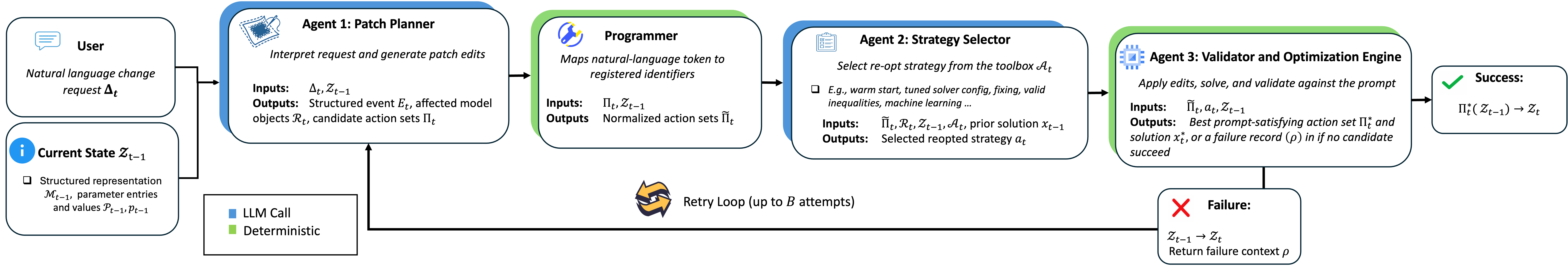} 
      \caption{\textbf{Zoom on Framework.} A bounded repair loop processes the user request $\Delta_t$ through three agents: the Patch Planner generates candidate edits, the Strategy Selector chooses a re-optimization strategy from the toolbox, and the Validator $+$ Optimization Engine applies the edits and solves. On validation failure, additional context $\rho$ is returned to Agent~1 (up to budget $B$). On success, the state advances to $\mathcal{Z}_t$, and the system handles the next request.}
      \label{fig:reopt_llm_framework}
  \end{figure}

The \textbf{ReOpt-LLM} framework processes each change $\Delta_t$ through three agent modules and one deterministic post-processor, illustrated in Figure~\ref{fig:reopt_llm_framework}. Agent~1, the integrated patch planner, jointly interprets the request and proposes candidate edits in a single LLM call. The deterministic programmer normalizes the result. Agent~2, the re-optimization strategy selector, picks a solve strategy from the toolbox. Agent~3, the validator and optimization engine, applies the edits and solves the updated model. A validation-triggered retry loop wraps the whole workflow.

\noindent \textbf{Agent 1: Patch Planner.}
The patch planner takes the natural-language description $\Delta_t$ together with a representation of the current state $\mathcal{Z}_{t-1}$ that includes the metadata $\mathrm{desc}(\cdot)$ and $\mathrm{tags}(\cdot)$ for every model component and named parameter entry. In a single LLM call, the planner returns the implemented structured-event fields corresponding to $E_t$, the relevant objects
\[
    \mathcal{R}_t \subseteq \mathcal{V}_{t-1} \cup \mathcal{C}_{t-1} \cup \mathcal{O}_{t-1} \cup \mathcal{P}_{t-1},
\]
and a collection of candidate action sets
\[
    \Pi_t = \{\Pi_t^{(1)}, \Pi_t^{(2)}, \ldots\},
\]
where each $\Pi_t^{(r)} = \{\pi_{r,1}, \pi_{r,2}, \ldots\}$ is an ordered group of patches intended to be applied together, with $r$ indexing candidate action sets. Combining interpretation and planning in one agent removes the extra interaction step between a separate classifier and a downstream patch generator: the planner conditions the proposed edit directly on $\Delta_t$, $\mathcal{R}_t$, and the metadata of the candidate target objects, which keeps the model's view of the request and the proposed action consistent. Working at the action-set level allows the planner to express edits that are only correct as a coordinated group, such as adding a new constraint family alongside an update to a related parameter.

\noindent \textbf{Programmer.}
Each candidate action set is normalized by a deterministic, problem-specific routine before it reaches the solver. Let $\widetilde{\Pi}_t = \texttt{Normalize}(\Pi_t, \mathcal{Z}_{t-1})$ denote the resulting collection of normalized action sets. The programmer canonicalizes entity labels (e.g., maps the natural-language token \texttt{``Plant 1''} to the registered identifier \texttt{P1}), reconciles index types between the planner's output and the model's internal index sets (e.g., promotes the list \texttt{[P2, C2]} to the tuple \texttt{(P2, C2)} expected by the \texttt{flows} family), and, when needed, rewrites operations into a more efficient or more numerically stable form (e.g., replacing an \texttt{UPDATE\_CONSTRAINT\_RHS} on the supply constraint with an equivalent \texttt{UPDATE\_PARAMETER} on the supply vector). The programmer is deterministic, validated against the structured model, and isolates the LLM-based planner from low-level naming and indexing details. Because the patch planner usually emits canonical operations directly, the programmer typically performs only minor index coercion; its role is most visible when a natural-language label or representation does not match the model's internal naming.

\noindent \textbf{Agent 2: Re-optimization Strategy Selector.}
Given the normalized action sets, the strategy selector chooses a re-optimization strategy from a toolbox of computational techniques. Let $\mathcal{A}_t$ denote the set of available re-optimization strategies at time $t$, which may include \emph{direct warm starts} from a saved base solution, \emph{heuristic warm starts} produced by a problem-specific construction, \emph{tuned solver configurations} obtained from offline tuning, \emph{variable-fixing rules}, \emph{valid inequalities}, and \emph{prior-solution} reuse. The selector maps the action sets, the affected components, previous-solution information, and the computational budget to a strategy
\[
    a_t = \texttt{SelectReoptStrategy}(\widetilde{\Pi}_t, \mathcal{R}_t, \mathcal{Z}_{t-1}, x_{t-1}, \mathcal{A}_t).
\]
Here $x_{t-1}$ denotes the saved solution associated with $\mathcal{Z}_{t-1}$, if available. This selector is separated from the patch planner because a correct model edit and an effective re-optimization strategy address different tasks: the former determines whether the requested model change has been expressed correctly, while the latter determines how the updated model should be solved within the time budget.

\noindent \textbf{Agent 3: Validator and Optimization Engine.}
For each normalized action set $\Pi \in \widetilde{\Pi}_t$ and selected strategy $a_t$, this module (1) applies the patch sequence $\mathcal{Z}^\Pi = (\mathcal{M}^\Pi, \mathcal{P}^\Pi, p^\Pi) = \Pi(\mathcal{Z}_{t-1})$, (2) regenerates the solver model from $\mathcal{M}^\Pi$ and $p^\Pi$, (3) configures the optimization run using $a_t$, (4) solves
\[
    x_t^{\Pi} \in \arg\min_{x \in \mathcal{X}(\mathcal{Z}^\Pi)} F(x;\mathcal{Z}^\Pi),
\]
and (5) returns feasibility, objective value, and diagnostics on constraint violations. These diagnostics also provide structured feedback to the end user regarding whether the generated modifications satisfy the intended operational request and preserve implementability under the updated conditions. Across the candidate action sets, the framework retains the action set with the lowest objective among the prompt-satisfying solutions.

\noindent \textbf{Validation-triggered retry.}
A failure at any stage, including unparseable planner output, patch-application errors, infeasible solves, or violations of prompt-specific constraints, is captured as a typed failure record and fed back to the patch planner as repair context. The planner is then re-invoked, conditioned on the failure record, and the resulting candidate action sets are processed by the programmer and Agents~2--3 again. This repair loop is bounded by a fixed retry budget (a small integer in the implementation). Retries are reported as reliability diagnostics, separating cases that succeed on the first attempt from those that require repair. This retry mechanism acts as a reliability layer that helps maintain robustness and correctness under imperfect natural-language interpretation.

\subsection{Closed-Loop Re-Optimization}\label{sec:5-closedloop}

At each time step $t$, the \textbf{ReOpt-LLM} framework executes the following procedure, with the inner repair loop bounded by a retry budget $B$:
\begin{enumerate}
    \item Initialize repair context $\rho \leftarrow \emptyset$ and attempt counter $n \leftarrow 0$.
    \item While $n < B$ and no candidate action set has succeeded:
        \begin{enumerate}
            \item $(E_t, \mathcal{R}_t, \Pi_t) \leftarrow \texttt{Plan}(\Delta_t, \mathcal{Z}_{t-1}, \rho)$ via the integrated patch planner.
            \item $\widetilde{\Pi}_t \leftarrow \texttt{Normalize}(\Pi_t, \mathcal{Z}_{t-1})$ via the programmer.
            \item Set $a_t \leftarrow \texttt{SelectReoptStrategy}(\widetilde{\Pi}_t, \mathcal{R}_t, \mathcal{Z}_{t-1}, x_{t-1}, \mathcal{A}_t)$.
            \item $(\Pi_t^\ast, x_t^\ast, \rho) \leftarrow \texttt{ValidateAndSolve}(\widetilde{\Pi}_t, a_t, \mathcal{Z}_{t-1})$, returning the best prompt-satisfying action set and solution, or a failure record in $\rho$ if no candidate succeeds.
            \item $n \leftarrow n + 1$.
        \end{enumerate}
    \item If no candidate action set succeeds, report $\rho$, set $\mathcal{Z}_t \leftarrow \mathcal{Z}_{t-1}$, and keep $x_t \leftarrow x_{t-1}$.
    \item Otherwise, set $\mathcal{Z}_t \leftarrow \Pi_t^\ast(\mathcal{Z}_{t-1})$ and store $x_t \leftarrow x_t^\ast$.
\end{enumerate}
The closed-loop procedure continually aligns the structured model representation with evolving operational conditions and user-specified rules, and the retry mechanism makes the framework robust to occasional planner errors.

The framework requires only (i) a structured state $\mathcal{Z}_t = (\mathcal{M}_t, \mathcal{P}_t, p_t)$, (ii) natural-language change descriptions $\Delta_t$, and (iii) access to an optimization solver. It therefore applies broadly to transportation and flow problems, production scheduling with vehicle assignment, exam timetabling, and other large-scale optimization problems that must be continuously adapted and re-optimized as operational conditions and decision requirements evolve.

\section{Toy Example}\label{section:6}

This section illustrates the framework on a small transportation instance. The example provides an interpretable execution trace showing how natural-language operational requests are translated into structured model edits, normalized, validated, and re-optimized through the full workflow. The resulting solutions are then compared against the baseline plan to confirm that the closed loop preserves feasibility, satisfies the intended operational request, and provides an interpretable trace of how the model was modified.

Consider a classical transportation problem where a set of plants must supply goods to a set of customers. Let $I$ denote the set of plants, indexed by $i$, and $J$ the set of customers, indexed by $j$. Each plant $i \in I$ has a limited supply $s_i$, and each customer $j \in J$ requires a demand quantity $d_j$. Shipping one unit of product from plant $i$ to customer $j$ incurs a transportation cost $\kappa_{ij}$. The decision variables $x_{ij}$ represent the quantity shipped from $i$ to $j$. The mathematical formulation is

\allowdisplaybreaks
\begin{align}
\min \quad & \sum_{i \in I} \sum_{j \in J} \kappa_{ij}\, x_{ij}
\label{obj:transport} \\[0.5em]
\text{s.t.} \quad
& \sum_{j \in J} x_{ij} \le s_i,
&& \forall\, i \in I,
\label{con:supply} \\[0.5em]
& \sum_{i \in I} x_{ij} \ge d_j,
&& \forall\, j \in J,
\label{con:demand} \\[0.5em]
& x_{ij} \ge 0,
&& \forall\, i \in I,\, j \in J.
\label{con:nonneg}
\end{align}

\noindent Objective~(\ref{obj:transport}) minimizes the total transportation cost incurred when shipping goods from plants to customers. Constraints~(\ref{con:supply}) enforce the supply capacity at each plant, ensuring that total outgoing shipments do not exceed available supply. Constraints~(\ref{con:demand}) guarantee that each customer's demand requirement is satisfied through inbound shipments. Finally, Constraints~(\ref{con:nonneg}) restrict all shipment quantities to be nonnegative. In the structured model representation of Section~\ref{sec:5-structured}, this problem has one variable family $\mathcal{V} = \{\mathtt{flows}\}$ with index set $I \times J$, two constraint families $\mathcal{C} = \{\mathtt{supply\_constraints}, \mathtt{demand\_constraints}\}$, and one objective component $\mathcal{O} = \{\mathtt{transport\_cost}\}$. The named parameter set is $\mathcal{P} = \{\mathtt{supply}, \mathtt{demand}, \mathtt{costs}\}$, with values collected in $p$ and descriptive metadata attached to each entry for retrieval.

The toy instance contains two plants and three customers. The supply capacities are $s_1 = 20$ and $s_2 = 45$; the customer demands are $d_1 = 12$, $d_2 = 15$, and $d_3 = 18$. Transportation costs follow the matrix
\[
\kappa =
\begin{pmatrix}
4 & 6 & 8 \\
5 & 4 & 3
\end{pmatrix}.
\]
The baseline transportation model achieves a minimum total cost of $162.0$. In this solution, Plant 1 supplies 12 units to Customer 1, while Plant 2 supplies 15 units to Customer 2 and 18 units to Customer 3. The allocation satisfies every demand and respects both supply capacities. Each prompt below is fed to the workflow starting from this baseline.

\subsection{Prompt 1: Plant Maintenance}

\noindent \textbf{User prompt.} \emph{``Plant 1 is going into urgent maintenance for the next two days, so it cannot ship anything.''}

\noindent \textbf{Patch Planner (Agent 1).} In a single LLM call, the planner emits the structured event with $S_t = \{\mathtt{plants}: [\mathtt{P1}]\}$ and edit summary \emph{``set supply of plant P1 to zero to represent urgent maintenance downtime''}, the relevant components $\mathcal{R}_t = \{\mathtt{supply}, \mathtt{supply\_constraints}, \mathtt{flows}\}$, and one candidate action set containing the single patch \texttt{UPDATE\_PARAMETER} on \texttt{supply} with key \texttt{P1} and value $0.0$. The two-day maintenance window is captured in the edit summary but not in the patch itself, since the toy model is single-period.

\noindent \textbf{Programmer.} The proposed patch is already canonical: the entity label \texttt{P1} matches the registered identifier and the parameter key is correctly typed. The programmer leaves the patch unchanged. Because the patch is explicit and localized, the end user can directly verify that the proposed edit correctly reflects the operational maintenance request before re-optimization.

\noindent \textbf{Strategy Selector (Agent 2).} The selector picks a warm strategy, with the recorded rationale that the edit sets one plant's supply to zero, expected reuse is high, and warm start is supported and available.

\noindent \textbf{Validator and Optimization Engine (Agent 3).} The patch is applied, the solver model is regenerated, and the solve completes feasibly. The total cost increases from $162$ to $174$, with all shipments rerouted through Plant 2.

\noindent The closed loop produces a feasible re-optimized plan that satisfies the user's maintenance request by eliminating all outbound shipments from Plant 1 while preserving demand feasibility. The cost increase reflects the loss of Plant 1's lower-cost capacity.

\subsection{Prompt 2: Transportation Shortage}

\noindent \textbf{User prompt.} \emph{``There is an unexpected shortage of trucks for deliveries from Plant 2 to Customer 2 this week. The maximum that can be shipped on this route is 5 units.''}

\noindent \textbf{Patch Planner (Agent 1).} The planner emits the structured event with $S_t = \{\mathtt{plants}: [\mathtt{P2}], \mathtt{customers}: [\mathtt{C2}]\}$ and edit summary \emph{``limit shipment flow from Plant 2 to Customer 2 to a maximum of 5 units due to truck shortage''}, relevant components $\mathcal{R}_t = \{\mathtt{flows}\}$, and one candidate action set containing the single patch \texttt{UPDATE\_BOUND} on \texttt{flows} with index \texttt{[P2, C2]}, bound type \texttt{upper}, and value $5.0$.

\noindent \textbf{Programmer.} The programmer promotes the index list \texttt{[P2, C2]} to the tuple \texttt{(P2, C2)} expected by the \texttt{flows} family. No structural rewrite is needed; the patch op remains \texttt{UPDATE\_BOUND}.

\noindent \textbf{Strategy Selector (Agent 2).} A warm strategy is selected, with the rationale that the route cap is a local change, expected reuse is high, and no structural modification is introduced.

\noindent \textbf{Validator and Optimization Engine (Agent 3).} The bound update is applied, and re-optimization redirects shipments from the capped route. The solve is feasible with total cost $184$, and the flow on $(\mathtt{P2}, \mathtt{C2})$ saturates at $5$; Plant 1 absorbs more of Customer 2's demand while Plant 2 covers the remaining Customer 1 demand needed to respect Plant 1's supply capacity.

\noindent 
The framework absorbs the route-level capacity restriction while satisfying the user's request that shipments on route $(\mathtt{P2}, \mathtt{C2})$ never exceed the imposed limit of $5$ units. It also does not modify any constraint family, illustrating that local capacity edits can be expressed as bound updates rather than as constraint-RHS edits.

\subsection{Prompt 3: Customer Order}

\noindent \textbf{User prompt.} \emph{``Customer 3 has placed an urgent order of 10 additional units on top of their normal demand.''}

\noindent \textbf{Patch Planner (Agent 1).} The planner emits the structured event with $S_t = \{\mathtt{customers}: [\mathtt{C3}]\}$ and edit summary \emph{``increase demand for customer C3 by 10 units due to urgent order''}, relevant components $\mathcal{R}_t = \{\mathtt{demand}, \mathtt{demand\_constraints}\}$, and one candidate action set containing the single patch \texttt{UPDATE\_PARAMETER} on \texttt{demand} with key \texttt{C3} and a delta of $+10$. By emitting the additive change as a \texttt{delta} field rather than an absolute value, the planner pushes the arithmetic onto the patch executor and avoids hand-computing the new total.

\noindent \textbf{Programmer.} The proposed patch is canonical and uses the additive form supported by the patch language. The programmer leaves the patch unchanged.

\noindent \textbf{Strategy Selector (Agent 2).} A warm strategy is selected, with the rationale that the demand increase is local, expected reuse is high, and no structural edit is introduced.

\noindent \textbf{Validator and Optimization Engine (Agent 3).} The patch is applied, increasing the demand at \texttt{C3} from $18$ to $28$, and the solve completes feasibly with total cost $192$. The increase reflects the additional shipments required to satisfy the urgent order.

\noindent The example illustrates how additive natural-language requests are absorbed at the parameter level via the \texttt{delta} update form while ensuring that the increased demand for Customer 3 is fully satisfied in the re-optimized solution. Letting the patch language carry the additive semantics keeps the patch planner free to operate at the component level and avoids placing arithmetic responsibilities on the LLM.

\section{Evaluation Protocol}\label{section:eval-protocol}

This section operationalizes the abstract LLM evaluation dimensions of Section~\ref{section:4} into the concrete experimental protocol used in both case studies. The framework variants, LLM models, design grid, retry mechanism, success criteria, validation procedures, and failure-mode taxonomy are common to the OCP and Cornell evaluations. Time limit, toolbox composition, and quality metric are deliberately case-specific because they reflect the managerial setting, and are described in the respective case-study experimental design subsections.

\subsection{Framework Variants}\label{sec:eval-variants}

The evaluation compares three variants of the framework, holding the rest of the pipeline fixed.

\noindent \textbf{\emph{ReOpt-LLM-Patch}.} The proposed implementation. The LLM expresses the user request through the structured patch language of Section~\ref{sec:5-patch}, the LLM selector chooses re-optimization techniques from the case-specific toolbox, and the validator executes the resulting model update.

\noindent \textbf{\emph{Direct-Code Agent}.} A baseline in which the LLM edits implementation code directly rather than using the structured patch language. This variant shares the validator and retry budget, but its outputs are arbitrary code edits rather than explicit, auditable, and reviewable patch operations.

\noindent \textbf{\emph{ReOpt-LLM-Patch without Selector}.} An ablation that retains the structured patch representation but removes the toolbox-selection step. The model update is applied and re-solved without any toolbox elements.

The toolbox composition itself is case-specific and is described in each case-study section.

\subsection{LLM Models, Design Grid, and Retry}\label{sec:eval-grid}

Each case study evaluates the three variants on three OpenAI models \citep{openai_api}: gpt-4.1-mini, gpt-4.1, and gpt-5. Each case provides five instances and six prompt classes, so the design contains $5\times 6\times 3\times 3 = 270$ LLM-assisted prompt-instance cases (instances $\times$ prompts $\times$ models $\times$ variants). Each model-by-variant row of an aggregate comparison table therefore aggregates 30 prompt-instance cases.

To support detailed instance-by-prompt reporting, each case study designates one configuration as its \emph{default} and reports a $5\times 6$ instance-by-prompt table for that configuration. The defaults are \emph{ReOpt-LLM-Patch} with gpt-4.1-mini for OCP and \emph{ReOpt-LLM-Patch} with gpt-5 for Cornell. All other configurations contribute only to the aggregate baseline and ablation comparisons.

All LLM-assisted runs share a validation-triggered retry mechanism with a uniform budget of one repair attempt. When the first attempt produces a patch that fails schema or semantic validation, the framework returns the validator's error feedback to the LLM and requests a single corrected attempt. This retry mechanism acts as a controlled reliability layer, allowing the framework to recover from imperfect natural-language interpretations while preserving validation consistency across all configurations.

\subsection{Success Criteria and Failure-Mode Taxonomy}\label{sec:eval-criteria}

A prompt-instance case is classified along the four nested success criteria in Table~\ref{tab:success_criteria} that evaluate whether the framework correctly interprets, validates, and operationally satisfies the user request. Each criterion implies the success of the criteria above it, so reported rates in aggregate comparison tables are non-increasing across the four criteria.

\begin{table}[!ht]
\centering
\small
\caption{Nested success criteria used in both case studies.}
\label{tab:success_criteria}
\begin{tabular}{lp{11cm}}
\toprule
\textbf{Criterion} & \textbf{Definition} \\
\midrule
Update correctness & The generated model edits correctly and transparently matches the intended operational update expressed in the prompt. The ground-truth update is reference-derived in each case study. \\
Prompt satisfaction & Update correctness, and additionally, a feasible recovered plan that satisfies the intended operational request while respecting all prompt-specific constraints. \\
First-attempt success & Update correctness and prompt satisfaction achieved on the first attempt. \\
Final success & Update correctness and prompt satisfaction achieved after the retry budget is applied. \\
\bottomrule
\end{tabular}
\end{table}

Failures are classified using the fixed taxonomy in Table~\ref{tab:failure_modes} to support consistent operational diagnosis, interpretability, and reliability analysis across both case studies. The modes are not mutually exclusive: a single case may exhibit several modes simultaneously.

\begin{table}[!ht]
\centering
\small
\caption{Failure-mode taxonomy used uniformly in both case studies.}
\label{tab:failure_modes}
\begin{tabular}{lp{11cm}}
\toprule
\textbf{Mode} & \textbf{Definition} \\
\midrule
\emph{Wrong comp.} & The LLM targeted the wrong model component. For direct-code runs this is folded into \emph{Bad update}, since the direct-code agent does not expose a structured component-targeting step. \\
\emph{Invalid patch} & The patch fails schema or application validation. \\
\emph{Bad update} & The applied update does not match the intended edit. \\
\emph{No incumbent} & The solver returned no feasible solution within the time limit (equivalently, no validated incumbent was accepted). \\
\emph{Prompt viol.} & A feasible incumbent was returned but its edit violates at least one prompt-specific operational constraint. \\
\emph{Missing output} & No usable validated framework output is returned. \\
\bottomrule
\end{tabular}
\end{table}

\subsection{Shared Computational Environment}\label{sec:eval-env}

Both case studies run on nodes with 2 Intel Xeon Gold 6226 processors at 2.7 GHz and 24 GB of memory per allocated node, using Python 3.11 with Gurobi/gurobipy 13.0.1. Reported solve time is wall-clock time measured inside the Gurobi optimization call. The number of CPU cores and threads, the per-case time limit, the toolbox composition, the quality metric, and the default LLM configuration are case-specific and are stated in the experimental design subsections of Sections~\ref{section:ocp} and~\ref{section:cornell}.

\section{OCP Group Case Study}\label{section:ocp}

This section presents the OCP case study, which focuses on large-scale downstream production and shipment scheduling in a complex real-world industrial supply chain. The section first summarizes the underlying optimization problem at the Jorf site, then describes the experimental design used to evaluate the proposed framework, reports the main computational results, and concludes with the resulting managerial insights.

\subsection{Problem Description}

The OCP case study focuses on downstream supply-chain operations at the Jorf site in Morocco, one of the world's largest phosphate-processing and export facilities. The optimization model is a large-scale mixed-integer linear program that integrates production, storage, and vessel-loading decisions over a monthly planning horizon with daily resolution. The model captures the physical flow of materials across a highly interconnected network of conveyors, pipelines, storage units, and quays, while accounting for a demand-driven order-fulfillment process and a large product portfolio. Key decisions include selecting shipments and loading schedules, developing daily production plans across multiple processing units, and determining the implied inventory and changeover trajectories. Given the scale and combinatorial complexity of the problem, the framework supports coordinated downstream scheduling and provides actionable insights for operational planning at a site responsible for a large share of the group's production. The original industrial study provides additional background on the model and operating context \citep{erraqabi5}.

\subsection{Experimental Design}

\noindent \textbf{Instances.} The OCP evaluation considers five realistic monthly instances: OCP$_1$, OCP$_2$, OCP$_{3}$, OCP$_{4}$, and OCP$_{5}$. Table~\ref{tab:ocp_instances_final} summarizes the season, planning horizon, number of vessels, total demand, and model size for each instance. In the computational results below, the fulfillment denominator counts the shipments that remain active after preprocessing for a given prompt-instance case, and may therefore differ slightly from the raw vessel count reported in the instance table.

\begin{table}[!ht]
\centering
\caption{OCP instances over different seasons.}
\label{tab:ocp_instances_final}
\resizebox{\textwidth}{!}{%
\begin{tabular}{llccrrrr}
\toprule
\textbf{ID} & \textbf{Season} & \textbf{Horizon} & \textbf{Vessels} & \textbf{Demand} & \textbf{Variables} & \textbf{Binaries} & \textbf{Constraints} \\
\midrule
\textbf{OCP$_1$} & Winter & 30 & 38 & 846{,}702 & 125{,}880 & 16{,}292 & 109{,}977 \\
\textbf{OCP$_2$} & Winter & 30 & 38 & 856{,}686 & 126{,}314 & 16{,}695 & 110{,}922 \\
\textbf{OCP$_{3}$} & Autumn & 32 & 62 & 2{,}031{,}400 & 947{,}598 & 17{,}966 & 695{,}693 \\
\textbf{OCP$_{4}$} & Summer & 24 & 40 & 1{,}043{,}330 & 298{,}693 & 33{,}284 & 235{,}344 \\
\textbf{OCP$_{5}$} & Autumn & 32 & 61 & 957{,}338 & 948{,}009 & 18{,}264 & 780{,}267 \\
\midrule
\multicolumn{2}{c}{\textbf{Avg}} & \textbf{29.6} & \textbf{47.8} & \textbf{1{,}147{,}091} & \textbf{489{,}299} & \textbf{20{,}500} & \textbf{386{,}441} \\
\bottomrule
\end{tabular}%
}
\end{table}

\noindent \textbf{Prompts.} The OCP prompt catalog contains six representative classes of online operational updates: demand adjustment, shipment arrival delay, weather disruption, production-capacity reduction, transportation-capacity degradation, and a composed prompt that combines demand and delay. These six prompt classes span demand-side, transportation-side, production-side, quay-side, and multi-update perturbations, and therefore provide a compact but representative testbed for online re-optimization. Table~\ref{tab:ocp_prompt_examples_final} gives one example for each prompt class.

\begin{table}[!ht]
\centering
\caption{Examples of user prompts for OCP re-optimization.}
\label{tab:ocp_prompt_examples_final}
\begin{tabular}{lp{14cm}}
\hline
\textbf{Prompt} & \textbf{Example} \\
\hline
$P_1$ & \emph{Shipment 13297 requested an increase of 1500 tonnes to its initial demand of 4000 tonnes.} \\
$P_2$ & \emph{The vessel corresponding to Shipment 13297 will have a delay and has a new expected arrival date of January 20th.} \\
$P_3$ & \emph{The weather is bad on January 12th, and no loading operations are maintained on that day.} \\
$P_4$ & \emph{Production Unit 0280 suffers from an unexpected breakdown and requires maintenance on the 17th, operating only 12 hours instead of 24.} \\
$P_5$ & \emph{The conveyors linking the storage point with Quai 1 will operate at 80\% of their nominal capacity.} \\
$P_6$ & \emph{Apply both updates together: Shipment 13297 requests an increase of 1500 tonnes, and the vessel corresponding to Shipment 13297 has a new expected arrival date of January 20th.} \\
\hline
\end{tabular}
\end{table}

\noindent \textbf{Re-optimization toolbox.} Following the evaluation protocol of Section~\ref{section:eval-protocol}, the detailed instance-by-instance results below use the OCP default configuration: \emph{ReOpt-LLM-Patch} with gpt-4.1-mini. The OCP toolbox is designed for online industrial re-optimization, where resilience and speed are both important \citep{wu2024towards}. Its main components are a history of solutions, a set of customized heuristics, and a set of instance-specific tuned Gurobi configurations. In the selector-enabled variant, the LLM chooses a legal combination of these toolbox elements before the modified model is re-solved.

\noindent \textbf{Computational environment.} In addition to the shared settings of Section~\ref{sec:eval-env}, the OCP solves used 8 CPU cores with Gurobi configured to use 8 threads. Each OCP re-optimization run was subject to a 300-second time limit to reflect the online decision-support setting.

\noindent \textbf{Validation and metrics.} Because the OCP setting is online, the primary performance criteria are correct model adaptation, feasible recovery, high shipment fulfillment, and fast runtime \citep{erraqabi1,erraqabi5}. Patch-based runs are validated against reference OCP edits using the success criteria and failure-mode taxonomy of Section~\ref{sec:eval-criteria}. The main reported metrics are fulfillment, solve time, and solver gap, where the gap is the solver-reported relative MIP gap for the internal OCP objective, not a gap in shipment fulfillment. Direct-code runs are assessed by feasible completion and the same metrics, but they do not expose the same structured semantic trace as the patch-based runs. 

\subsection{Computation Results}

\noindent \textbf{Headline performance.} Table~\ref{tab:ocp_default_case_results_final} reports the case-by-case results for the default \emph{ReOpt-LLM-Patch} configuration on the five OCP instances and six prompt classes. Across the full $5\times 6$ evaluation grid, the framework returns a feasible incumbent for all 30 cases. The generated patch matches the reference edit on all 30 cases; prompt-satisfying recovery is achieved on 29 of 30. The only prompt-level failure is OCP$_{5}$--$P_1$: the demand-update patch is applied correctly, but re-optimization under the 300-second limit returns only 7 of 58 active shipments (versus reference recovery with substantially higher fulfillment). That incumbent is reported for diagnostic completeness. Among prompt-satisfying cases, the most difficult recovery is OCP$_{5}$--$P_4$, which fulfills 49 of 58 active shipments with a 25.96\% solver gap. These results show that the framework is operationally robust across the evaluation grid, while severe perturbations can still reduce the number of active shipments that can be fulfilled within the time cap.

\begin{table}[!ht]
\centering
\scriptsize
\caption{Case-by-case OCP results for the default \emph{ReOpt-LLM-Patch} configuration. \emph{Fulfilled} reports fulfilled shipments over the active post-preprocessing denominator for that prompt-instance case.}
\label{tab:ocp_default_case_results_final}
\resizebox{\textwidth}{!}{%
\begin{tabular}{llccrrr}
\toprule
\textbf{Instance} & \textbf{Prompt} & \textbf{Update correct} & \textbf{Prompt satisfied} & \textbf{Fulfilled} & \textbf{Time (s)} & \textbf{Gap (\%)} \\
\midrule
\textbf{OCP$_1$} & P1 & \checkmark & \checkmark & 37/38 & 37 & 0.00\% \\
\textbf{OCP$_1$} & P2 & \checkmark & \checkmark & 38/38 & 4 & 0.00\% \\
\textbf{OCP$_1$} & P3 & \checkmark & \checkmark & 37/38 & 300 & 3.67\% \\
\textbf{OCP$_1$} & P4 & \checkmark & \checkmark & 38/38 & 49 & 0.00\% \\
\textbf{OCP$_1$} & P5 & \checkmark & \checkmark & 38/38 & 53 & 0.00\% \\
\textbf{OCP$_1$} & P6 & \checkmark & \checkmark & 37/38 & 36 & 0.00\% \\
\addlinespace
\textbf{OCP$_2$} & P1 & \checkmark & \checkmark & 37/38 & 107 & 0.00\% \\
\textbf{OCP$_2$} & P2 & \checkmark & \checkmark & 38/38 & 3 & 0.00\% \\
\textbf{OCP$_2$} & P3 & \checkmark & \checkmark & 38/38 & 110 & 0.00\% \\
\textbf{OCP$_2$} & P4 & \checkmark & \checkmark & 38/38 & 107 & 0.00\% \\
\textbf{OCP$_2$} & P5 & \checkmark & \checkmark & 38/38 & 160 & 0.00\% \\
\textbf{OCP$_2$} & P6 & \checkmark & \checkmark & 37/38 & 107 & 0.00\% \\
\addlinespace
\textbf{OCP$_{3}$} & P1 & \checkmark & \checkmark & 59/60 & 98 & 0.00\% \\
\textbf{OCP$_{3}$} & P2 & \checkmark & \checkmark & 60/60 & 131 & 0.00\% \\
\textbf{OCP$_{3}$} & P3 & \checkmark & \checkmark & 60/60 & 152 & 0.00\% \\
\textbf{OCP$_{3}$} & P4 & \checkmark & \checkmark & 60/60 & 12 & 0.00\% \\
\textbf{OCP$_{3}$} & P5 & \checkmark & \checkmark & 60/60 & 12 & 0.00\% \\
\textbf{OCP$_{3}$} & P6 & \checkmark & \checkmark & 59/60 & 99 & 0.00\% \\
\addlinespace
\textbf{OCP$_{4}$} & P1 & \checkmark & \checkmark & 37/38 & 85 & 0.00\% \\
\textbf{OCP$_{4}$} & P2 & \checkmark & \checkmark & 38/38 & 112 & 0.00\% \\
\textbf{OCP$_{4}$} & P3 & \checkmark & \checkmark & 38/38 & 100 & 0.00\% \\
\textbf{OCP$_{4}$} & P4 & \checkmark & \checkmark & 38/38 & 4 & 0.42\% \\
\textbf{OCP$_{4}$} & P5 & \checkmark & \checkmark & 38/38 & 4 & 0.42\% \\
\textbf{OCP$_{4}$} & P6 & \checkmark & \checkmark & 37/38 & 89 & 0.21\% \\
\addlinespace
\textbf{OCP$_{5}$} & P1 & \checkmark & $\times$ & 7/58 & 300 & 3.65\% \\
\textbf{OCP$_{5}$} & P2 & \checkmark & \checkmark & 58/58 & 14 & 0.00\% \\
\textbf{OCP$_{5}$} & P3 & \checkmark & \checkmark & 55/58 & 300 & 3.96\% \\
\textbf{OCP$_{5}$} & P4 & \checkmark & \checkmark & 49/58 & 300 & 25.96\% \\
\textbf{OCP$_{5}$} & P5 & \checkmark & \checkmark & 58/58 & 13 & 0.00\% \\
\textbf{OCP$_{5}$} & P6 & \checkmark & \checkmark & 58/58 & 18 & 0.00\% \\
\bottomrule
\end{tabular}%
}
\end{table}

\noindent \textbf{Baseline comparison.} Table~\ref{tab:ocp_patch_vs_code_final} compares the structured patch formulation with direct code editing across the three LLMs. In this setting, \emph{update correctness} requires that the generated model edit match the ground-truth OCP update, while \emph{prompt satisfaction} additionally requires a feasible recovered plan. Since no configuration attains an additional semantic success through retry, first-attempt and final success coincide in this comparison. The structured patch representation is consistently more reliable than direct code editing: \emph{ReOpt-LLM-Patch} attains final success rates of 96.7\%, 93.3\%, and 93.3\% for gpt-4.1-mini, gpt-4.1, and gpt-5, respectively, whereas the direct-code baseline does not produce a semantically correct update in any of the evaluated cases. In these cases, the direct-code agent often produces executable or partially executable modifications, but the resulting model does not match the reference OCP edit semantics, so the solver is frequently applied to the wrong modified problem. These results indicate that, for OCP, the main advantage of the patch interface is semantic reliability rather than feasibility alone.

\begin{table}[!ht]
\centering
\scriptsize
\caption{Baseline comparison between the \emph{Direct-Code Agent} and \emph{ReOpt-LLM-Patch} in the OCP evaluation. Each row aggregates 30 prompt-instance cases. Entries are percentages and follow the success criteria of Section~\ref{sec:eval-criteria}.}
\label{tab:ocp_patch_vs_code_final}
\resizebox{\textwidth}{!}{%
\begin{tabular}{llrrrr}
\toprule
\textbf{Model} & \textbf{Method} & \textbf{Update correctness} & \textbf{Prompt satisfaction} & \textbf{First-attempt success} & \textbf{Final success} \\
\midrule
gpt-4.1-mini & \emph{Direct-Code Agent} & 0.0 & 0.0 & 0.0 & 0.0 \\
gpt-4.1-mini & \emph{ReOpt-LLM-Patch} & 100.0 & 96.7 & 96.7 & 96.7 \\
\addlinespace
gpt-4.1 & \emph{Direct-Code Agent} & 0.0 & 0.0 & 0.0 & 0.0 \\
gpt-4.1 & \emph{ReOpt-LLM-Patch} & 100.0 & 96.7 & 96.7 & 96.7 \\
\addlinespace
gpt-5 & \emph{Direct-Code Agent} & 0.0 & 0.0 & 0.0 & 0.0 \\
gpt-5 & \emph{ReOpt-LLM-Patch} & 100.0 & 96.7 & 96.7 & 96.7 \\
\bottomrule
\end{tabular}%
}
\end{table}

\noindent \textbf{Failure analysis.} Table~\ref{tab:ocp_failure_modes_final} summarizes the failure patterns underlying the method comparison. For patch-based runs, failures are sparse and concentrated in a small number of cases, most notably a recurring demand-update miss on OCP$_{5}$--$P_1$, which appears as an invalid patch application. By contrast, the direct-code baseline more often returns solver-feasible plans whose edits do not match the intended ground-truth update. As a result, the direct-code rows are dominated by \emph{bad update} and prompt-level semantic violations rather than by pure optimization failure. This distinction explains why the direct-code baseline may still recover incumbents with nontrivial fulfillment while failing to achieve semantic success.

\begin{table}[!ht]
\centering
\scriptsize
\caption{Failure modes for the \emph{Direct-Code Agent} and \emph{ReOpt-LLM-Patch} in the OCP evaluation. \emph{Note.} Each entry counts prompt-instance cases (out of 30) that exhibit the corresponding failure mode. Columns follow the taxonomy of Table~\ref{tab:failure_modes}.}
\label{tab:ocp_failure_modes_final}
\resizebox{\textwidth}{!}{%
\begin{tabular}{llrrrrrr}
\toprule
\textbf{Model} & \textbf{Variant} & \textbf{Wrong comp.} & \textbf{Invalid patch} & \textbf{Bad update} & \textbf{No incumbent} & \textbf{Prompt viol.} & \textbf{Missing output} \\
\midrule
gpt-4.1-mini & \emph{Direct-Code Agent} & 0 & 0 & 30 & 3 & 27 & 3 \\
gpt-4.1-mini & \emph{ReOpt-LLM-Patch} & 1 & 1 & 1 & 1 & 0 & 1 \\
\addlinespace
gpt-4.1 & \emph{Direct-Code Agent} & 0 & 0 & 30 & 5 & 25 & 5 \\
gpt-4.1 & \emph{ReOpt-LLM-Patch} & 2 & 1 & 2 & 2 & 0 & 2 \\
\addlinespace
gpt-5 & \emph{Direct-Code Agent} & 0 & 0 & 30 & 1 & 29 & 1 \\
gpt-5 & \emph{ReOpt-LLM-Patch} & 2 & 1 & 2 & 2 & 0 & 2 \\
\bottomrule
\end{tabular}%
}
\end{table}

\noindent \textbf{Selector ablation.} Table~\ref{tab:ocp_selector_ablation_final} shows the effect of the LLM-guided toolbox selector for the default OCP configuration. The table reports feasible completion rather than semantic success. Under that operational criterion, both variants return an incumbent on all 30 cases, but selector guidance materially improves operational performance: mean fulfillment increases from 86.57\% to 95.65\%, mean runtime falls from 194.88 seconds to 97.20 seconds, and mean solver gap decreases from 2.24\% to 1.28\%. Mean fulfillment is reported as an unweighted average of the 30 case-level fulfillment percentages. Among the 29 semantically valid recoveries, mean fulfillment increases further to 98.54\%, indicating that the low overall mean is driven primarily by the single rejected OCP$_{5}$--$P_1$ diagnostic run. In the online OCP setting, the main value of toolbox selection is therefore not binary solve success alone, but faster recovery and higher-quality shipment plans under tight time limits.

\begin{table}[!ht]
\centering
\scriptsize
\caption{Ablation of the LLM-guided re-optimization toolbox selector for the default OCP configuration. Lower runtime and lower gap are better.}
\label{tab:ocp_selector_ablation_final}
\resizebox{\textwidth}{!}{%
\begin{tabular}{lrrrrrr}
\toprule
\textbf{Variant} & \textbf{Feasible comp.} & \textbf{No incumbent} & \textbf{Invalid / build fail.} & \textbf{Mean fulfill. (\%)} & \textbf{Mean time (s)} & \textbf{Mean gap} \\
\midrule
ReOpt-LLM-Patch (selector), gpt-4.1-mini & 100.0\% & 0 & 0 & 95.65\% & 97.20 & 1.28\% \\
ReOpt-LLM-Patch without selector, gpt-4.1-mini & 100.0\% & 0 & 0 & 86.57\% & 194.88 & 2.24\% \\
\bottomrule
\end{tabular}%
}
\end{table}

\noindent \textbf{Operational-quality diagnostics.} The semantic failure table above explains why the structured patch interface is necessary, but it does not replace the online quality story. In the default configuration, OCP$_{5}$--$P_1$ is the lone semantic failure, while OCP$_{5}$--$P_4$ is the hardest semantically valid operational case, with fulfillment dropping to 49/58 and the solver gap rising to 25.96\%. These cases motivate the use of fulfillment, runtime, and gap as the main online evaluation criteria in addition to semantic correctness.

\subsection{Managerial Insights}

The OCP case highlights the managerial value of LLM-assisted re-optimization in large-scale industrial environments subject to continuous operational disruptions. In practice, planners must rapidly react to demand changes, vessel delays, bad weather, unit breakdowns, or transportation-capacity degradation while preserving as much of the accepted operational plan as possible. The managerial challenge is therefore not only solving a large-scale optimization model, but continuously adapting the deployed decision-support system as business conditions evolve. The relevant benchmark is thus whether the system can quickly interpret operational disruptions, correctly update the optimization model, recover feasible shipment plans, and maintain high fulfillment under strict operational time limits.

The results show that the structured patch-based framework provides this operational reliability. In the default OCP configuration, \emph{ReOpt-LLM-Patch} returns a feasible incumbent on all thirty prompt-instance pairs, achieves semantically valid prompt-satisfying recovery on 29 of the 30 cases, and maintains 95.65\% mean fulfillment across all raw incumbents. Among the 29 semantically valid recoveries, mean fulfillment increases to 98.54\%. Importantly, planners can express disruptions directly in natural language while obtaining validated and auditable model updates without requiring continuous intervention from the optimization engineering team. This is managerially significant because many deployed optimization systems gradually lose effectiveness as operational rules evolve faster than models can be manually maintained.

The structured patch representation is particularly important from a governance and implementation perspective because it creates a transparent and traceable link between the user request, the affected model components, the applied optimization edits, and the final solution. This improves interpretability, facilitates validation, and increases organizational trust in AI-assisted optimization systems.

The toolbox selector provides a complementary operational benefit. Even when multiple variants recover feasible plans, selector-guided re-optimization produces higher fulfillment, shorter runtimes, and lower solver gaps under the same online time budget. This distinction matters operationally because a technically feasible plan with poor fulfillment or slow recovery may still be unacceptable in practice. More broadly, the case illustrates how LLM-assisted re-optimization can reduce decision latency in complex industrial environments by transforming natural-language disruption reports into validated optimization edits and rapid re-solves.

\section{Cornell University Case Study}\label{section:cornell}

This section presents the Cornell University exam scheduling case study, illustrating the application of the \textbf{ReOpt-LLM} framework to a large-scale real-world academic scheduling problem. It first describes the exam scheduling optimization model and its key constraints, then outlines the experimental design, presents the computational results, and concludes with managerial insights relevant to university administrators.

\subsection{Problem Description}

The second case study is drawn from the final exam scheduling problem at Cornell University, where an integer programming framework was developed to address the complex and highly constrained nature of university-wide exam timetabling. The model accommodates a wide range of institutional requirements, including the front-loading of large courses, the exclusion of specific time slots, and policies governing exam conflicts. By generating and comparing multiple model variants and incorporating heuristic solution approaches, the framework supports informed decision-making by the university registrar, allowing explicit trade-offs between schedule quality and student conflicts. The approach demonstrated clear improvements over traditional lecture time-based scheduling methods, yielding substantial administrative time savings and increased satisfaction among students and faculty \citep{ye2025cornell}.

The evaluation focuses on the block-sequencing stage of the Cornell framework. Exam blocks are treated as already constructed by the block-assignment stage, and the re-optimization task assigns these fixed blocks to exam time slots under updated policy and schedule-quality requirements. The block-sequencing MIP enforces assignment and continuity constraints so that each exam block is placed once and each slot receives a valid block sequence. It also exposes policy constraint families that are natural targets for re-optimization. Front-loading constraints require large exam blocks to be scheduled before an instance-specific cutoff slot. Reserved-slot requests are handled through virtual blocks. Fixing a virtual block to a reserved slot keeps that slot unavailable to real exam blocks while preserving the one-block-per-slot sequencing structure. Day-level capacity constraints can be added by aggregating block enrollments over the slots in a given day. The objective penalizes weighted counts of undesirable student exam patterns, including same-day triples, triples within 24 hours, evening-to-morning back-to-back exams, other back-to-back exams, and three exams in four consecutive slots.

This case differs from the OCP setting in two important ways. First, re-optimization is offline, i.e., the registrar can evaluate alternative schedules before publication. Consequently, the revised schedule need not remain close to a previously implemented plan. Second, given that the re-optimization is offline, the primary managerial objective is schedule quality rather than speed. The evaluation, therefore, assesses \textbf{ReOpt-LLM} by whether it correctly interprets scheduling-policy prompts, applies the intended model update, satisfies the resulting constraints, and produces schedules whose objective values are close to a saved hindsight reference incumbent obtained under the same time limit.

\subsection{Experimental Design}

\noindent \textbf{Instances.} The evaluation uses five calibrated synthetic Cornell exam-scheduling instances generated with the \href{https://github.com/Joeyetinghan/exam-scheduling-mip-generator}{Cornell final-exam scheduling MIP generator}. The generator builds block-assignment and block-sequencing MIPs using anonymized co-enrollment, triplet co-enrollment, and exam-size inputs, and provides a synthetic data generator for larger calibrated instances. The formulations evaluated here are block-sequencing MIPs, with the block assignment taken as fixed. Table~\ref{tab:exam_instances} reports the paper-facing instance identifiers together with the semester label, number of exams, available time slots, exam blocks, binary variables, and constraints used for each instance. All variables in these generated formulations are binary.

\begin{table}[!t]
\centering
\caption{Cornell exam-scheduling instances.}
\label{tab:exam_instances}
\resizebox{\textwidth}{!}{%
\begin{tabular}{llrrrrr}
\toprule
\textbf{Instance} & \textbf{Semester} & \textbf{\# Exams} & \textbf{\# Slots} & \textbf{\# Blocks} & \textbf{\# Binary} & \textbf{\# Constraints} \\
\midrule
\textbf{EXAM$_1$} & Spring 2024 & 544 & 24 & 20 & 677{,}376 & 401{,}011 \\
\textbf{EXAM$_2$} & Fall 2023 & 601 & 24 & 18 & 677{,}376 & 401{,}010 \\
\textbf{EXAM$_3$} & Spring 2023 & 553 & 25 & 17 & 796{,}875 & 468{,}866 \\
\textbf{EXAM$_4$} & Fall 2022 & 588 & 24 & 19 & 677{,}376 & 401{,}010 \\
\textbf{EXAM$_5$} & Spring 2022 & 539 & 24 & 16 & 677{,}376 & 401{,}006 \\
\midrule
\multicolumn{2}{c}{\textbf{Avg}} & \textbf{565} & \textbf{24} & \textbf{18} & \textbf{701{,}276} & \textbf{414{,}581} \\
\bottomrule
\end{tabular}
}
\end{table}

\noindent \textbf{Prompts.} The exam prompt catalog contains six representative prompt classes that capture policy changes and late operational updates faced during exam scheduling. These prompts cover reserved time slots, late co-enrollment changes, large-exam deadlines, stress-related objective-weight adjustments, day-level proctoring capacity, and a composed update that applies multiple policy changes in sequence. Table~\ref{tab:prompt_examples_exam} gives one example for each prompt class. The large-exam deadline in prompt $P_3$ is instance-specific. Large exams must be completed before slot 20 for EXAM$_1$, slot 19 for EXAM$_2$, slot 17 for EXAM$_3$, slot 19 for EXAM$_4$, and slot 15 for EXAM$_5$. Prompt $P_6$ exercises the composition of three updates that target disjoint parts of the model: $P_4$ modifies objective weights, $P_2$ modifies co-enrollment parameters, and $P_1$ modifies slot availability. The composed model is invariant to the order in which these updates are applied, so the listed sequence is illustrative. 

\begin{table}[!t]
\centering
\caption{Examples of user prompts for Cornell exam-scheduling re-optimization.}
\label{tab:prompt_examples_exam}
\begin{tabular}{lp{14cm}}
\hline
\textbf{Prompt} & \textbf{Example} \\
\hline
$P_1$ & \emph{Reserve the evening slot immediately before the final evening slot so the staff can begin arranging the auditorium for graduation events.} \\
$P_2$ & \emph{Increase the pairwise co-enrollment count between Block 4 and Block 9 by 120 students due to late add/drop changes.} \\
$P_3$ & \emph{Ensure all large exams with over 300 students are completed before the instance-specific cutoff time slot to allow teaching assistants sufficient grading time.} \\
$P_4$ & \emph{The Student Assembly raised concerns about extreme stress; increase the penalty for having ``three exams in 24 hours'' to be 20 times that of a regular back-to-back.} \\
$P_5$ & \emph{Due to an unexpected shortage of available proctors, limit the total number of students taking exams on Day 2 to a maximum of 4,000.} \\
$P_6$ & \emph{A combination of prompts, e.g., $P_4$, then $P_2$, then $P_1$.} \\
\hline
\end{tabular}
\end{table}

\noindent \textbf{Re-optimization toolbox.} Following the evaluation protocol of Section~\ref{section:eval-protocol}, the detailed instance-by-prompt and schedule-quality results below use the Cornell default configuration: \emph{ReOpt-LLM-Patch} with gpt-5. The exam-scheduling toolbox consists of three reusable elements that can be applied individually or in combination. \emph{Direct warm start} uses the saved base block-sequencing solution as a Gurobi MIP start for the modified model. \emph{Heuristic warm start} builds a greedy schedule after the prompt-specific edit by first fixing the virtual-block assignments required for reserved slots and then placing constrained and low-enrollment blocks to respect front-loading and day-load-cap restrictions. The resulting schedule is converted to Gurobi variable starts. \emph{Tuned configuration} applies instance-specific Gurobi parameter files obtained from the tuning workflow. In \emph{ReOpt-LLM-Patch}, the LLM selector chooses a feasible executable combination of these elements for each prompt-instance case, such as \emph{direct+heuristic+tuned}. The scratch label is not a toolbox element. It is used only in \emph{ReOpt-LLM-Patch without Selector} to disable the selector and force a solve without any toolbox elements.

\noindent \textbf{Computational environment.} In addition to the shared settings of Section~\ref{sec:eval-env}, the Cornell solves used 6 CPU cores with Gurobi configured to use 6 threads. Each Cornell re-optimization run was subject to a 3600-second time limit to reflect the offline pre-publication setting.

\noindent \textbf{Validation and metrics.} Because the Cornell setting is offline, the primary performance criterion is schedule quality at publication time. Patch-based runs are validated against deterministic prompt-specific reference-script edits using the success criteria and failure-mode taxonomy of Section~\ref{sec:eval-criteria}. The main reported metrics are the objective difference $\Delta \mathrm{obj}=\mathrm{obj}_{\mathrm{LLM}}-\mathrm{obj}_{\mathrm{Ref}}$ and the corresponding reference-relative percentage gap against a saved hindsight reference incumbent solved with the \emph{direct+heuristic+tuned} configuration under the same 3600-second time limit. A negative reference-relative gap on a valid, prompt-satisfying case means the framework found a better incumbent than the hindsight reference, while negative gaps from invalid configurations are not interpreted as quality improvements. Schedule quality is additionally reported using the metrics of \citet{ye2025cornell}: triples, back-to-back exams, two exams in 24 hours, and three exams in four slots. Direct conflicts, also defined in \citet{ye2025cornell}, are zero in every evaluated case and are omitted from the reported tables.

\subsection{Computation Results}

\noindent \textbf{Headline performance.} Table~\ref{tab:cornell_main_results} reports the instance-by-prompt results for the default \emph{ReOpt-LLM-Patch} configuration. The framework satisfies all 30 prompts and produces no prompt-specific constraint violations. Nineteen of the 30 prompt-instance cases match the saved hindsight reference incumbent exactly. The largest reference-relative deviations occur for prompt classes $P_2$ and $P_4$, which modify the co-enrollment structure and objective weights, respectively. These cases are expected to be more sensitive to the interaction between the re-optimization technique and the solver's incumbent search under the 3600-second cap.

\begin{table}[!ht]
\centering
\small
\caption{Cornell exam-scheduling results for the default \emph{ReOpt-LLM-Patch} configuration. The objective difference $\Delta\mathrm{obj}=\mathrm{ReOpt\ obj.}-\mathrm{Ref.\ obj.}$ is reported so that exact matches with the hindsight reference incumbent ($\Delta\mathrm{obj}=0$) can be read directly. Reference-relative percentage gaps are visualized in Figure~\ref{fig:cornell_reference_gap}, and per-prompt schedule-quality metrics in Table~\ref{tab:cornell_schedule_quality}.}
\label{tab:cornell_main_results}
\begin{tabular}{llccrrr}
\toprule
\textbf{Instance} & \textbf{Prompt} & \textbf{Update correct} & \textbf{Prompt satisfied} & \textbf{ReOpt obj.} & \textbf{Ref. obj.} & \textbf{$\Delta$obj} \\
\midrule
EXAM$_1$ & P1 & \checkmark & \checkmark & 6{,}901 & 6{,}901 & 0 \\
EXAM$_1$ & P2 & \checkmark & \checkmark & 6{,}662 & 5{,}338 & 1{,}324 \\
EXAM$_1$ & P3 & \checkmark & \checkmark & 8{,}615 & 8{,}615 & 0 \\
EXAM$_1$ & P4 & \checkmark & \checkmark & 7{,}756 & 5{,}938 & 1{,}818 \\
EXAM$_1$ & P5 & \checkmark & \checkmark & 8{,}871 & 8{,}871 & 0 \\
EXAM$_1$ & P6 & \checkmark & \checkmark & 9{,}325 & 9{,}325 & 0 \\
\addlinespace
EXAM$_2$ & P1 & \checkmark & \checkmark & 7{,}336 & 7{,}336 & 0 \\
EXAM$_2$ & P2 & \checkmark & \checkmark & 6{,}331 & 6{,}331 & 0 \\
EXAM$_2$ & P3 & \checkmark & \checkmark & 9{,}719 & 9{,}576 & 143 \\
EXAM$_2$ & P4 & \checkmark & \checkmark & 6{,}834 & 6{,}847 & -13 \\
EXAM$_2$ & P5 & \checkmark & \checkmark & 6{,}410 & 6{,}410 & 0 \\
EXAM$_2$ & P6 & \checkmark & \checkmark & 8{,}306 & 8{,}204 & 102 \\
\addlinespace
EXAM$_3$ & P1 & \checkmark & \checkmark & 4{,}915 & 4{,}915 & 0 \\
EXAM$_3$ & P2 & \checkmark & \checkmark & 4{,}003 & 4{,}003 & 0 \\
EXAM$_3$ & P3 & \checkmark & \checkmark & 9{,}468 & 9{,}468 & 0 \\
EXAM$_3$ & P4 & \checkmark & \checkmark & 4{,}233 & 4{,}233 & 0 \\
EXAM$_3$ & P5 & \checkmark & \checkmark & 4{,}150 & 4{,}150 & 0 \\
EXAM$_3$ & P6 & \checkmark & \checkmark & 5{,}409 & 5{,}409 & 0 \\
\addlinespace
EXAM$_4$ & P1 & \checkmark & \checkmark & 6{,}488 & 6{,}488 & 0 \\
EXAM$_4$ & P2 & \checkmark & \checkmark & 5{,}648 & 5{,}648 & 0 \\
EXAM$_4$ & P3 & \checkmark & \checkmark & 8{,}452 & 8{,}306 & 146 \\
EXAM$_4$ & P4 & \checkmark & \checkmark & 6{,}215 & 6{,}194 & 21 \\
EXAM$_4$ & P5 & \checkmark & \checkmark & 5{,}849 & 5{,}779 & 70 \\
EXAM$_4$ & P6 & \checkmark & \checkmark & 7{,}376 & 7{,}376 & 0 \\
\addlinespace
EXAM$_5$ & P1 & \checkmark & \checkmark & 2{,}645 & 2{,}645 & 0 \\
EXAM$_5$ & P2 & \checkmark & \checkmark & 2{,}560 & 2{,}537 & 23 \\
EXAM$_5$ & P3 & \checkmark & \checkmark & 9{,}237 & 9{,}237 & 0 \\
EXAM$_5$ & P4 & \checkmark & \checkmark & 2{,}534 & 2{,}528 & 6 \\
EXAM$_5$ & P5 & \checkmark & \checkmark & 2{,}570 & 2{,}570 & 0 \\
EXAM$_5$ & P6 & \checkmark & \checkmark & 2{,}674 & 2{,}674 & 0 \\
\bottomrule
\end{tabular}
\end{table}

\begin{figure}[!ht]
    \centering
    \includegraphics[width=0.88\linewidth]{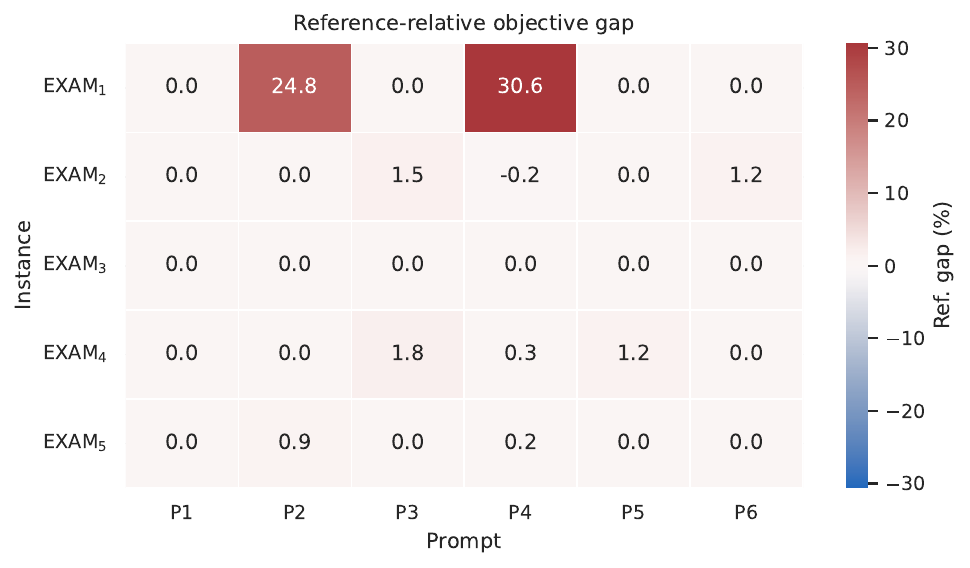}
    \caption{Reference-relative objective gap for the default \emph{ReOpt-LLM-Patch} configuration.}
    \label{fig:cornell_reference_gap}
\end{figure}

Figure~\ref{fig:cornell_reference_gap} visualizes the reference-relative objective gaps for these prompt-instance cases. Most cases are at or near zero, while the few larger deviations are concentrated in the more structurally consequential prompts. The figure reinforces the main conclusion from Table~\ref{tab:cornell_main_results}. \emph{ReOpt-LLM-Patch} consistently produces valid, prompt-satisfying schedules and usually remains close to the saved hindsight reference incumbent under the same time budget.

\noindent \textbf{Schedule-quality details.} Table~\ref{tab:cornell_schedule_quality} aggregates the schedule-quality metrics by prompt and compares them with the saved 3600-second hindsight reference incumbents. The aggregate deltas are small relative to the scale of the instances.

\begin{table}[!ht]
\centering
\scriptsize
\caption{Schedule-quality comparison for the default \emph{ReOpt-LLM-Patch} configuration. Each cell aggregates the metric across the five Cornell instances. \emph{ReOpt} is the LLM-assisted result, \emph{Ref.} is the saved hindsight reference incumbent, and $\Delta$ is their difference.}
\label{tab:cornell_schedule_quality}
\resizebox{\textwidth}{!}{%
\begin{tabular}{llrrrrrrr}
\toprule
\textbf{Metric} & \textbf{Series} & \textbf{Overall} & \textbf{P1} & \textbf{P2} & \textbf{P3} & \textbf{P4} & \textbf{P5} & \textbf{P6} \\
\midrule
Triples & ReOpt & 126.5 & 109.8 & 88.8 & 237.8 & 89.4 & 118.2 & 115.0 \\
Triples & Ref. & 120.6 & 109.8 & 77.2 & 233.6 & 73.6 & 115.8 & 113.8 \\
Triples & $\Delta$ & 5.9 & 0 & 11.6 & 4.2 & 15.8 & 2.4 & 1.2 \\
\addlinespace
Back-to-back & ReOpt & 3{,}338.9 & 3{,}267.0 & 3{,}028.8 & 4{,}229.2 & 3{,}044.4 & 3{,}119.0 & 3{,}345.0 \\
Back-to-back & Ref. & 3{,}319.7 & 3{,}267.0 & 2{,}974.4 & 4{,}226.4 & 2{,}987.0 & 3{,}120.0 & 3{,}343.6 \\
Back-to-back & $\Delta$ & 19.2 & 0 & 54.4 & 2.8 & 57.4 & -1.0 & 1.4 \\
\addlinespace
Two in 24 hours & ReOpt & 10{,}399.6 & 10{,}245.0 & 9{,}892.4 & 12{,}167.2 & 9{,}742.6 & 9{,}916.8 & 10{,}433.6 \\
Two in 24 hours & Ref. & 10{,}383.7 & 10{,}245.0 & 9{,}843.8 & 12{,}166.0 & 9{,}620.4 & 9{,}989.0 & 10{,}438.2 \\
Two in 24 hours & $\Delta$ & 15.9 & 0 & 48.6 & 1.2 & 122.2 & -72.2 & -4.6 \\
\addlinespace
Three in four slots & ReOpt & 603.5 & 549.0 & 484.2 & 1{,}001.6 & 466.8 & 517.4 & 601.8 \\
Three in four slots & Ref. & 593.0 & 549.0 & 446.8 & 994.4 & 446.2 & 520.8 & 600.8 \\
Three in four slots & $\Delta$ & 10.5 & 0 & 37.4 & 7.2 & 20.6 & -3.4 & 1.0 \\
\bottomrule
\end{tabular}%
}
\end{table}

\noindent \textbf{Baseline comparison.} Table~\ref{tab:cornell_patch_vs_code} compares \emph{ReOpt-LLM-Patch} with the \emph{Direct-Code Agent}. For each model, the patch-based implementation has a higher final success rate than direct code editing: 63.3\% versus 3.3\% for gpt-4.1-mini, 83.3\% versus 26.7\% for gpt-4.1, and 100\% versus 33.3\% for gpt-5. The model comparison also shows that stronger LLMs improve reliability, but model strength alone does not remove the need for the structured framework. The \emph{Direct-Code Agent} remains below 50\% final success even with gpt-5, indicating that direct implementation-code editing introduces grounding and validation challenges that are not resolved by simply using a stronger model.

\begin{table}[!ht]
\centering
\scriptsize
\caption{Baseline comparison between the \emph{Direct-Code Agent} and \emph{ReOpt-LLM-Patch} in the Cornell evaluation. \emph{Note.} Each row aggregates 30 prompt-instance cases and entries are the percentage that satisfy the corresponding criterion. The four nested success criteria are defined in Section~\ref{sec:eval-criteria}.}
\label{tab:cornell_patch_vs_code}
\resizebox{\textwidth}{!}{%
\begin{tabular}{llrrrr}
\toprule
\textbf{Model} & \textbf{Method} & \textbf{Update correctness} & \textbf{Prompt satisfaction} & \textbf{First-attempt success} & \textbf{Final success} \\
\midrule
gpt-4.1-mini & \emph{Direct-Code Agent} & 13.3 & 3.3 & 0.0 & 3.3 \\
gpt-4.1-mini & \emph{ReOpt-LLM-Patch} & 63.3 & 63.3 & 60.0 & 63.3 \\
\addlinespace
gpt-4.1 & \emph{Direct-Code Agent} & 30.0 & 26.7 & 26.7 & 26.7 \\
gpt-4.1 & \emph{ReOpt-LLM-Patch} & 83.3 & 83.3 & 83.3 & 83.3 \\
\addlinespace
gpt-5 & \emph{Direct-Code Agent} & 36.7 & 33.3 & 26.7 & 33.3 \\
gpt-5 & \emph{ReOpt-LLM-Patch} & 100.0 & 100.0 & 100.0 & 100.0 \\
\bottomrule
\end{tabular}%
}
\end{table}

\noindent \textbf{Retry diagnostics.} The first-attempt and final success columns of Table~\ref{tab:cornell_patch_vs_code} characterize the retry mechanism's contribution. The default \emph{ReOpt-LLM-Patch} configuration with gpt-5 succeeds on the first attempt throughout the evaluation and does not require retries. For weaker configurations, the retry mechanism provides a small reliability layer: gpt-4.1-mini \emph{ReOpt-LLM-Patch} improves from 60.0\% to 63.3\%, gpt-5 \emph{Direct-Code Agent} from 26.7\% to 33.3\%, and gpt-4.1-mini \emph{Direct-Code Agent} from 0.0\% to 3.3\% after retries are applied.

\noindent \textbf{Failure analysis.} Table~\ref{tab:cornell_failure_modes} summarizes the failure modes behind the direct-code and structured-patch comparison. Direct code editing most often fails because code changes cannot be grounded cleanly back to the intended scheduling operation, or because the resulting update satisfies only part of the prompt. For example, the \emph{Direct-Code Agent} has 25, 21, and 19 prompt-specific violations for gpt-4.1-mini, gpt-4.1, and gpt-5, respectively. The structured patch interface reduces these failures by forcing the LLM output into auditable model-edit operations before optimization.

\begin{table}[!ht]
\centering
\scriptsize
\caption{Failure modes for the \emph{Direct-Code Agent} and \emph{ReOpt-LLM-Patch} in the Cornell evaluation. \emph{Note.} Each entry counts prompt-instance cases (out of 30) that exhibit the corresponding failure mode. Columns follow the taxonomy of Table~\ref{tab:failure_modes}. General-violation and retry-exhausted counts are zero throughout and are omitted.}
\label{tab:cornell_failure_modes}
\resizebox{\textwidth}{!}{%
\begin{tabular}{llrrrrrr}
\toprule
\textbf{Model} & \textbf{Variant} & \textbf{Wrong comp.} & \textbf{Invalid patch} & \textbf{Bad update} & \textbf{No incumbent} & \textbf{Prompt viol.} & \textbf{Missing output} \\
\midrule
gpt-4.1-mini & \emph{Direct-Code Agent} & 2 & 1 & 26 & 2 & 25 & 2 \\
gpt-4.1-mini & \emph{ReOpt-LLM-Patch} & 2 & 0 & 11 & 0 & 11 & 0 \\
\addlinespace
gpt-4.1 & \emph{Direct-Code Agent} & 0 & 0 & 21 & 0 & 21 & 0 \\
gpt-4.1 & \emph{ReOpt-LLM-Patch} & 0 & 0 & 5 & 0 & 5 & 0 \\
\addlinespace
gpt-5 & \emph{Direct-Code Agent} & 5 & 3 & 19 & 5 & 19 & 5 \\
gpt-5 & \emph{ReOpt-LLM-Patch} & 0 & 0 & 0 & 0 & 0 & 0 \\
\bottomrule
\end{tabular}%
}
\end{table}

\noindent \textbf{Selector ablation.} Table~\ref{tab:cornell_selector_ablation} compares \emph{ReOpt-LLM-Patch} with \emph{ReOpt-LLM-Patch without Selector} for the default exam-scheduling configuration. The ablation preserves the structured patch plan but forces a solve without any toolbox elements or their combinations. Holding the model-update representation fixed, LLM-guided toolbox selection raises final success from 83.3\% to 100\%, eliminates five no-incumbent cases, and reduces the median objective difference from 1{,}441 to 0 (median reference-relative gap from 25.7\% to 0.0\%) under the same 3600-second time limit. The mean objective difference falls from 3{,}577{,}316.3 to 121.3, but is dominated by four no-selector cases in which the time limit expired with the solver still in a high-triple-penalty incumbent region. This pattern supports a two-part interpretation of the framework. Patch editing improves semantic control over the model update. In contrast, toolbox selection improves the reliability and incumbent quality of the subsequent re-optimization.

\begin{table}[!ht]
\centering
\scriptsize
\caption{Effect of the LLM-guided toolbox selector on the default exam-scheduling configuration. Each row aggregates 30 prompt-instance cases for gpt-5. Objective and gap summaries exclude no-incumbent cases. Lower values are better in every column.}
\label{tab:cornell_selector_ablation}
\resizebox{\textwidth}{!}{%
\begin{tabular}{lrrrrrr}
\toprule
\textbf{Variant} & \textbf{Final success} & \textbf{No incumbent} & \textbf{Mean $\Delta$obj} & \textbf{Median $\Delta$obj} & \textbf{Mean ref. gap} & \textbf{Median ref. gap} \\
\midrule
\emph{ReOpt-LLM-Patch} & 100.0\% & 0 & 121.3 & 0.0 & 2.1\% & 0.0\% \\
\emph{ReOpt-LLM-Patch without Selector} & 83.3\% & 5 & 3{,}577{,}316.3 & 1{,}441.0 & 88{,}459.9\% & 25.7\% \\
\bottomrule
\end{tabular}%
}
\end{table}

Table~\ref{tab:cornell_selector_ablation_paired} extends the comparison across all three models on prompt-instance cases where both variants produced prompt-satisfying incumbents: toolbox selection produces a lower objective on 13 of 16 cases for gpt-4.1-mini, 16 of 20 for gpt-4.1, and 22 of 25 for gpt-5, with median objective improvements above 1{,}400 in every model.

\begin{table}[!ht]
\centering
\scriptsize
\caption{Paired objective improvement from LLM-guided toolbox selection in the Cornell evaluation. Each row restricts attention to prompt-instance cases for which both \emph{ReOpt-LLM-Patch} and \emph{ReOpt-LLM-Patch without Selector} produced prompt-satisfying incumbents. Objective improvement is no-selector minus selector. Positive values favor the selector variant.}
\label{tab:cornell_selector_ablation_paired}
\resizebox{\textwidth}{!}{%
\begin{tabular}{lrrrr}
\toprule
\textbf{Model} & \textbf{Common valid cases} & \textbf{Selector lower obj.} & \textbf{Median obj. improvement} & \textbf{Mean obj. improvement} \\
\midrule
gpt-4.1-mini & 16 & 13/16 & 1{,}430.5 & 5{,}588{,}645.4 \\
gpt-4.1 & 20 & 16/20 & 1{,}430.5 & 4{,}471{,}130.5 \\
gpt-5 & 25 & 22/25 & 1{,}441.0 & 3{,}577{,}173.0 \\
\bottomrule
\end{tabular}%
}
\end{table}

\subsection{Managerial Insights}

The Cornell case highlights the value of LLM-assisted re-optimization in settings where the organization must repeatedly adjust an unpublished plan before committing to it. Before the final exam schedule is released, the registrar can revise the model in response to late policy changes, stakeholder concerns, or capacity information without requiring a full return to the original OR modeling team. This shifts the operational workflow from a sequential, expert-dependent process to an interactive and decentralized decision process in which end users can directly adapt the model. The relevant standard is therefore not proximity to the previous schedule, but whether the modified model is auditable, feasible, and capable of producing a high-quality schedule before publication. In this offline setting, decision quality dominates speed, and the ability to explore and validate multiple alternatives before commitment becomes the primary managerial objective.

A key implication is the reduction in decision latency and the expansion of decision-making capabilities. In traditional settings, incorporating new constraints or policy updates may require multiple iterations between stakeholders and OR experts, often taking days. The proposed framework enables the registrar to evaluate multiple what-if scenarios within the same decision cycle, effectively transforming re-optimization into an interactive planning tool rather than a batch process. This capability is particularly valuable in academic scheduling, where late updates are frequent and where evaluating alternative policies, such as stress reduction or capacity adjustments, is essential before finalizing the schedule.

The structured patch interface is central to this managerial value. It creates a traceable chain from the natural-language prompt to the affected model component, the model update, the validation checks, and the final schedule. This audit trail is especially important for academic scheduling, where students, faculty, or administrators may contest policy changes. Beyond transparency, this traceability serves as a governance mechanism, reducing the risk of unintended or inconsistent model modifications and facilitating communication across stakeholders. The results show that the default \emph{ReOpt-LLM-Patch} configuration satisfies all tested prompt classes, produces no prompt-specific violations, and remains close to the saved hindsight reference incumbent on average.

More broadly, the case illustrates how LLM-assisted re-optimization reduces dependence on OR experts while preserving solution reliability, thereby enabling a separation between policy specification and model implementation. The retry mechanism provides an additional reliability layer for weaker configurations, although the default configuration succeeds on the first attempt throughout the evaluation. These insights extend beyond academic scheduling to other pre-commitment planning settings, where organizations must iteratively refine decisions before execution, highlighting the potential of LLM-based systems to enhance adaptability, accountability, and decision quality in complex operational environments.

\section{Conclusions}\label{section:conclusions}

This paper proposes an LLM-assisted collaborative re-optimization framework, \textbf{ReOpt-LLM}, that integrates end users and optimization to enable adaptive decision support. The framework formalizes how user prompts, reflecting new rules or operational changes, translate into structured and auditable updates of the core optimization model, enabling the system to evolve dynamically with changing environments. By embedding this process within a language interface, the framework lowers the barrier between domain users and large-scale mathematical optimization models while preserving solver-based validation and mathematical rigor. Methodologically, the paper highlights that large-scale re-optimization is fundamentally a structured model-reasoning problem rather than a mere re-solving task. The proposed framework combines interpretable patch-based model edits with solver-aware re-optimization strategies driven by a toolbox of OR techniques, including historical solutions, valid inequalities, solver configurations, heuristics, and ML-guided components. The computational results further show that structured model-edit interfaces and toolbox-aware orchestration significantly improve reliability and solution quality compared with direct code-editing approaches. Managerially, the framework enables organizations to sustain and adapt deployed optimization systems without constant expert intervention, thereby improving the long-term sustainability of decision-support tools in dynamic operational settings. Future work may extend this foundation to stochastic, robust, or multiobjective settings and evaluate its performance across additional large-scale applications. Ultimately, the framework advances the vision of sustainable, human-centered optimization systems that remain adaptive, interpretable, and operationally effective under continuous change.

\section*{Acknowledgments}

This research was partly supported by the NSF AI Institute for Advances in Optimization (Award 2112533).
  
\begin{appendices}

\section{Framework Artifacts}\label{app:framework}

This appendix documents the agent prompts, the patch JSON schema, the validator decision logic, and the direct-code baseline used in both case studies.

\subsection{Agent System Prompts}\label{app:prompts}

The pipeline uses three LLM agents in sequence: a Patch Planner, a Strategy Selector, and an optional Code-Edit Planner used only by the direct-code baseline of Appendix~\ref{app:codeagent}.

\subsubsection{Patch Planner}

The Patch Planner converts a user prompt $\Delta_t$ into a candidate set of structured patches. Its system instruction is reproduced below.

\begin{promptbox}[title=Patch Planner -- system instruction]
You are a reoptimization planner. Use the deterministic model representation to interpret the requested change and propose candidate model edits. Return JSON only. Use \texttt{candidate\_action\_sets} as the canonical output format.

Each \texttt{candidate\_action\_set} is one executable plan; put all coordinated edits for a single plan in the same action set. Each item must be a JSON object with \texttt{actions=[...]}; the patch list key inside each item must be \texttt{actions}. Each patch object must use the canonical keys \texttt{op}, \texttt{target}, \texttt{scope}, \texttt{update}, and optional \texttt{notes}. Return multiple candidate action sets only when they are genuinely different alternative plans, and do not duplicate the same edits in both grouped and flat forms.

Patch payloads must be executable as written: use concrete ids and numeric literals for indices, row labels, and values whenever the model representation provides enough information. Do not emit pseudo-code, formulas, set names, or symbolic placeholders inside patch indices or values. For keyed parameter edits, place the concrete sub-index in \texttt{update.key}. For numeric requests phrased as ``increase by'', ``decrease by'', or other additive changes, use \texttt{update.delta} instead of overwriting with \texttt{update.value}. For \texttt{materialized\_linear} constraint families, use matching concrete row ids in \texttt{lhs\_spec.rows} and \texttt{rhs\_spec}, and concrete executable variable indices in every term. If the representation explicitly exposes a compact problem-specific semantic \texttt{lhs\_spec.kind}, that semantic payload may be used instead of materializing every row term. If a valid executable patch cannot be expressed, return empty candidate lists rather than a symbolic or guessed placeholder patch.

Required JSON keys:
\texttt{edit\_summary} (short free-form summary of the requested edit);
\texttt{affected\_sets} (object mapping entity or set labels to identifiers mentioned or strongly implied by the delta, or \texttt{\{\}} if none);
\texttt{relevant\_components} (list of model component names);
\texttt{candidate\_action\_sets} (list of executable candidate plans, each \texttt{\{actions: [\dots]\}});
\texttt{planning\_hints} (optional planner hints such as \texttt{edit\_scope='local|structural'} or \texttt{expected\_reuse='high|low'}).
\end{promptbox}

At runtime, the framework appends to this system message a per-case block of problem-specific guidance (Appendices~\ref{app:framing-ocp} and~\ref{app:framing-cornell}), the list of allowed patch operators for the case, generic operator guidance, and the patch schemas of Table~\ref{tab:patch_ops}. On validation failure, a structured repair-context block is added to the user message of the next planning call, prefaced by the prelude reproduced below.

\begin{promptbox}[title=Patch Planner -- repair-context prelude (retry only)]
This is a fresh repair attempt for the same user request from a fresh planning pass. Preserve the user's intent, not the previous implementation details. Use the runtime feedback below only to avoid the previous failure mode. The items below are runtime feedback from earlier attempts in this same run.
\end{promptbox}

The repair-context block itself consists of bullet entries derived from the fields \texttt{failure\_stage}, \texttt{failure\_kind}, \texttt{failure\_message}, \texttt{repair\_instruction}, and an \texttt{attempt\_history} list of recent failures, under the one-attempt retry budget of Section~\ref{sec:eval-grid}.

\subsubsection{Strategy Selector}

The Strategy Selector reads the validated patch set and the case-specific toolbox catalog, then emits a JSON object with the required keys \texttt{solve\_strategy}, \texttt{toolbox\_plan}, \texttt{rationale}, and the optional key \texttt{confidence} $\in[0,1]$.

\begin{promptbox}[title=Strategy Selector -- system instruction]
You choose the fastest safe reoptimization solve strategy. Return JSON only.

Pick exactly one solve strategy from the allowed list. Do not invent toolbox items or unsupported strategies. Toolbox plans are executable in this runtime and must match the chosen solve strategy.

Prefer \texttt{warm+tuned} over \texttt{warm} alone when both warm reuse and tuned solving are available and the edit looks reuse-friendly. Prefer warm reuse for local edits when a reusable solution exists but tuned solving is unavailable or unnecessary. Prefer tuned or scratch for structural edits when warm reuse looks fragile.

Required JSON keys: \texttt{solve\_strategy}, \texttt{toolbox\_plan}, \texttt{rationale}. Optional JSON key: \texttt{confidence} as a number in $[0,1]$.
\end{promptbox}

\subsubsection{OCP Case-Specific Framing}\label{app:framing-ocp}

The OCP setup injects a problem-context block into every Patch-Planner call. The block describes the OCP supply-chain model (its objective, variable, and constraint naming conventions, and date encoding) so the planner can interpret prompts and emit regex-driven patches against the pre-compiled LP. Unlike the Cornell setup, OCP does not inject a separate patch-form guidance block. The three OCP-applicable operations (the pattern-based family in Table~\ref{tab:patch_ops}) and their schemas are surfaced through the generic operator-guidance mechanism.

\begin{promptbox}[title=OCP domain framing -- problem context]
You are working with a large-scale mixed-integer linear program (MILP) for the OCP Group downstream supply-chain at the Jorf site in Morocco. The model integrates production, storage, and vessel-loading decisions over a monthly planning horizon with daily resolution. The model is provided as a pre-compiled LP file; you do NOT have access to the Python source code that generated it. All modifications must be expressed as structured patches that operate on variable and constraint names via regex patterns.

\textbf{Objective.} The model maximizes total shipment fulfillment (\texttt{TotalFulfillment}). Secondary objectives (changeover minimization, lateness penalties, etc.) are present but weighted at 0 in the base formulation.

\textbf{Variable naming conventions.} All variable names end with a suffix of the form \texttt{\_(<numeric\_id>)\#<integer>}. These suffixes are instance-specific and must be matched via regex, never hard-coded. Use the prefix (everything before the first parenthesis) for pattern matching. 

\textbf{Constraint naming conventions.} Constraint names also end with instance-specific suffixes. The match is by prefix.

\end{promptbox}

\subsubsection{Cornell Case-Specific Framing}\label{app:framing-cornell}

The Cornell setup injects two complementary domain blocks into every Patch-Planner call. The first describes the exam-scheduling model, its calendar, objective, core variables, and grounding data, so the planner can interpret prompts in the Registrar's vocabulary. The second pins each prompt class to a canonical exam patch form, preventing the planner from inventing alternative constraint-family names for behaviors that the exam model already supports.

\begin{promptbox}[title=Cornell domain framing -- problem context]
You are helping the University Registrar evaluate changes to the final-exam schedule. This is the block-sequencing stage of a Group-then-Sequence workflow: exams have already been grouped into blocks, and this model places those blocks into exam slots. The model is about exam timing only; room assignment is handled separately. Delta requests usually describe policy, comfort, or operational changes to the exam calendar.

\textbf{Calendar and basic assumptions.} The exam period is a fixed ordered sequence of slots. The standard interpretation is three slots per day -- morning, afternoon, and evening, typically \texttt{9am}, \texttt{2pm}, and \texttt{7pm}. Requests like ``Day 2'', ``morning'', or ``the evening slot immediately before the final evening slot'' should be grounded through \texttt{slots\_per\_day}, \texttt{slot\_times}, and slot ids, not through guessed LP names. Some slots may be intentionally excluded from use; in final patch proposals, relative calendar phrases should be resolved to explicit slot ids when the instance data makes that possible.

\textbf{Objective.} The sequencing objective penalizes stressful student exam patterns: \texttt{alpha} (triples within one day), \texttt{beta} (triples within 24 hours), \texttt{gamma1} (evening-to-morning back-to-backs), \texttt{gamma2} (other back-to-backs), and \texttt{delta} (three exams in four consecutive slots). These events are not double-counted: exam pairs already part of a triple are not also counted as back-to-backs.

\textbf{Core model view.} The sequencing formulation is cyclic over the slot set. \texttt{x[i,j,k,s]=1} means block~$i$ is placed at slot~$s$, block~$j$ at slot~$s+1$, and block~$k$ at slot~$s+2$; assignment-like policy rules are expressed by summing \texttt{x[i,*,*,s]} terms over the relevant slots. The variables \texttt{y[...]} and \texttt{z[...]} are linkage variables used to score triple and four-slot patterns. Assignment and continuity constraints enforce a valid cyclic schedule in which each block is placed once and each slot receives one block.

\textbf{Grounding data.} The rendered model representation already exposes \texttt{virtual\_blocks}, \texttt{large\_blocks}, \texttt{early\_slots}, \texttt{reserved\_slots}, \texttt{block\_enrollment}, \texttt{pair\_counts}, \texttt{triplet\_counts}, \texttt{slots\_per\_day}, and \texttt{slot\_times}. Use those rendered values directly instead of inventing symbolic placeholders, and use the exact numeric slot cutoff stated in the prompt when front-loading requests are instance-specific.

\textbf{Canonical interpretations.} Slot-reservation requests use \texttt{reserved\_virtual\_slot}; front-loading requests are interpreted through \texttt{large\_blocks} and \texttt{early\_slots}; day-load requests use \texttt{block\_enrollment} and the first-index occupancy view of \texttt{x[i,j,k,s]}; co-enrollment changes usually affect \texttt{pair\_counts} or \texttt{triplet\_counts}, where pairwise block relationships are unordered, so both ordered pair keys are updated unless the request explicitly distinguishes direction; weight or comfort tradeoff requests usually affect \texttt{alpha}, \texttt{beta}, \texttt{gamma1}, \texttt{gamma2}, or \texttt{delta}. When the instance data resolves a slot or day directly, explicit slot ids are emitted rather than symbolic formulas.

\textbf{Combined requests.} The requested order is preserved when one request bundles several changes; combined Registrar edits may mix weight changes, co-enrollment edits, front-loading, slot reservations, and day-load restrictions.
\end{promptbox}

\begin{promptbox}[title=Cornell domain framing -- patch-form guidance]
Use the canonical exam patch forms. Do not invent new constraint-family names or semantic \texttt{lhs} kinds when an existing exam family fits.

For slot exclusion requests like the penultimate evening slot, emit \texttt{ADD\_CONSTRAINT\_FAMILY} with \texttt{constraint.name='reserved\_virtual\_slot'}, \texttt{lhs\_spec.kind='reserved\_virtual\_slot'}, the explicit grounded slot id, and the smallest available virtual block id.

For large-exam frontloading requests, emit \texttt{UPDATE\_PARAMETER(name='early\_slots', value=[\dots])} with explicit slot ids; do not invent a new frontload family name.

For pairwise co-enrollment changes between two blocks, emit keyed \texttt{UPDATE\_PARAMETER} patches for both ordered pair keys \texttt{[a,b]} and \texttt{[b,a]} on the pair-count override surface unless the request explicitly distinguishes direction.

For day-level load caps, emit \texttt{ADD\_CONSTRAINT\_FAMILY} with \texttt{constraint.name='slot\_load\_cap'}, \texttt{lhs\_spec.kind='slot\_load\_cap'}, explicit slot ids, and a stable row id such as \texttt{'day\_2'}.
\end{promptbox}

\subsection{Validator Decision Logic}\label{app:validator}

The validator is a best-improvement candidate solver. For each candidate patch in the planner's output:
\begin{enumerate}
\item Apply the patch to a copy of the base model. Catch schema or application errors and record them as a failure entry. On success, proceed.
\item Invoke the Gurobi solver under the case-specific time limit, providing a warm start when the Strategy Selector has specified one.
\item Distinguish between (i) a clean infeasible or no-incumbent solve and (ii) unexpected errors, which surface the original exception.
\item Compare returned incumbents across surviving candidates and retain the one with the lowest objective value.
\end{enumerate}
If all candidates fail, the validator returns the structured feedback consumed by the retry mechanism. The validator does not perform a separate semantic check on the LLM's intent. Semantic correctness is judged by whether the executed patch produces a feasible incumbent that satisfies the prompt-specific constraints encoded by the Update-correctness criterion of Section~\ref{sec:eval-criteria}.

\subsection{Direct-Code Agent Specification}\label{app:codeagent}

The direct-code baseline delegates code editing to the open-source Aider CLI \citep{aider}. Its system instruction is reproduced below.

\begin{promptbox}[title=Direct-Code Agent -- system instruction]
You are editing a packaged re-optimization problem through Aider. Modify only the editable files provided to you; use any read-only files only as context. Keep the code executable and minimal. Prefer changing the canonical solver implementation that is rebuilt and re-solved. Honor the natural-language request and any generic problem guidance provided below.
\end{promptbox}

When the surfaced editing surface is an LP-wrapper, as in the OCP case, the system instruction is extended with two LP-wrapper rules.

\begin{promptbox}[title=Direct-Code Agent -- LP-wrapper extension]
For LP-wrapper editing, \texttt{build\_codeedit\_model(...)} must only mutate and return a Gurobi model. Do not solve the model or inspect solution values inside \texttt{build\_codeedit\_model(...)}.
\end{promptbox}

A case-specific code-edit guidance block is appended for both case studies, keeping the agent inside the existing solver surface rather than rewriting model structure.

\begin{promptbox}[title=Direct-Code Agent -- OCP code-edit guidance]
For LP-backed codeedit runs, edit only \texttt{lp\_codeedit\_wrapper.py}. Keep the wrapper thin and generic, operate on actual LP variable and constraint names via regex, and do not invent new \texttt{runtime\_data} keys or instance-specific helper APIs.
\end{promptbox}

\begin{promptbox}[title=Direct-Code Agent -- Cornell code-edit guidance]
Use the existing exam solver mechanisms rather than inventing new interfaces or rewriting the model structure. For slot exclusion requests, resolve the concrete slot from \texttt{slot\_times} and \texttt{slots\_per\_day}, update \texttt{reserved\_slots}, and preserve the virtual-block reservation mechanism that materializes the slot exclusion. For pairwise co-enrollment changes between two blocks, keep pair-count updates symmetric across both ordered keys unless the request explicitly distinguishes direction. For large-exam frontloading requests, reuse \texttt{early\_slots} and the existing large-blocks logic instead of adding a new frontload formulation. For day-level load caps, ground the limit with \texttt{block\_enrollment} over the affected day slots and preserve the existing slot-load-cap semantics. For stress-penalty changes, update the existing objective weight parameters rather than rewriting the objective structure.
\end{promptbox}

The agent's action space is the set of unified diffs that Aider can apply to the surfaced editable files, and its output is the captured unified diff together with the run status. The same one-attempt retry budget is applied: a failed compilation or solve raises the validator's feedback back to Aider, which produces a corrected diff. Aider is invoked with the same OpenAI model under evaluation, so the only protocol-level difference between \emph{ReOpt-LLM-Patch} and \emph{Direct-Code Agent} is the action representation -- a structured patch versus an arbitrary code edit.

\section{Methodology Details}\label{app:method}

\subsection{Gurobi Tuning Workflow}\label{app:tuning}

Both case studies use Gurobi's built-in tuning tool to produce instance-specific parameter files. Tuning is performed once per instance on the \emph{base} (unperturbed) model, before any prompt is applied. The same tuned parameter file is then reused across all prompts, framework variants, and LLM models for that instance, including the LLM-assisted runs and the hindsight reference run. This ensures that no variant or prompt sees a parameter file tuned for its specific perturbation, so the tuned configuration acts as a fixed instance-level baseline rather than as a per-case advantage.

The time budgets differ between the two case studies. For OCP, the Gurobi \texttt{TuneTimeLimit} is set to 28{,}800 seconds (8 hours) and each tuning candidate is given a 1{,}800-second (30-minute) solve budget before \texttt{model.tune()} is invoked. For Cornell, the \texttt{TuneTimeLimit} is set to 86{,}400 seconds (24 hours), with no per-candidate solve cap set explicitly.

\subsection{OCP Re-optimization Heuristics}\label{app:ocp-heuristics}

The OCP re-optimization heuristics are inspired by the rollback-and-repair strategy of \cite{wu2024towards}, and are designed to generate high-quality warm starts from a previously optimized solution under localized disruptions. The general principle is to preserve as much of the incumbent solution structure as possible while selectively relaxing decisions that are directly affected by the prompt-induced changes. This ensures fast recovery and strong initial incumbents for the subsequent MIP solve.

At a high level, the heuristic proceeds in two steps. First, the solver retrieves the set of decision variables that are active in the base solution, which represents a complete feasible plan before disruption. Second, it applies prompt-dependent filtering rules that identify which parts of the solution are invalidated by the new operational context (e.g., demand changes, delays, capacity reductions, weather disruptions, or production constraints). Variables unaffected by the disruption are fixed to their incumbent values to preserve solution structure, while those directly impacted are released to allow re-optimization. The resulting partially fixed solution is then passed as a warm start to Gurobi, enabling rapid recovery of feasibility and improved incumbent quality under the new constraints.

\subsection{Exam-Scheduling Heuristic Warm Start}\label{app:exam-warmstart}

The exam-scheduling heuristic warm start constructs a feasible block-to-slot assignment that respects prompt-induced policy changes before passing it to Gurobi as a MIP start. The construction proceeds in four stages, summarized in Algorithm~\ref{alg:exam-warmstart}: reserved-slot pinning, front-loading of large exams, day-load capping, and a default fallback to the cached base assignment.

The notation is as follows. $\mathcal{B}$ is the block set with enrollment function $e:\mathcal{B}\to\mathbb{N}$. $\mathcal{S}$ is the slot set, partitioned by exam day as $\mathcal{S}=\bigsqcup_{d\in\mathcal{D}}\mathcal{S}_d$, where $\mathcal{D}$ is the set of exam days. The prompt parameter tuple $\theta=(\theta_\mathrm{rsv},\tau,c,\theta_\mathrm{cap})$ specifies a partial reserved-slot map $\theta_\mathrm{rsv}$ for virtual blocks, a large-exam enrollment threshold $\tau$, a front-loading cutoff slot $c$, and a partial day-enrollment cap $\theta_\mathrm{cap}$. The cached base assignment from the prior solve is $X^0:\mathcal{B}\to\mathcal{S}$. The algorithm builds the output $X$ incrementally, using a free-slot pool $F\subseteq\mathcal{S}$ that tracks the slots not yet assigned to any block. A block is \emph{unassigned} if it is not in $\mathrm{dom}(X)$. Ties in any sort are broken by block id in ascending order.

\begin{algorithm}[!ht]
\SetAlgoLined
\DontPrintSemicolon
\KwIn{Block set $\mathcal{B}$, slot set $\mathcal{S}=\bigsqcup_{d\in\mathcal{D}}\mathcal{S}_d$, parameters $\theta=(\theta_\mathrm{rsv},\tau,c,\theta_\mathrm{cap})$, base assignment $X^0:\mathcal{B}\to\mathcal{S}$.}
\KwOut{Warm-start assignment $X:\mathcal{B}\to\mathcal{S}$ supplied to Gurobi as a MIP start.}
\BlankLine
$X \leftarrow \emptyset$;\quad $F \leftarrow \mathcal{S}$ \tcp*{$F$ is the free-slot pool}
\BlankLine
\tcp{Stage 1: pin reserved slots}
\ForEach{$v \in \mathrm{dom}(\theta_\mathrm{rsv})$}{
    $s \leftarrow \theta_\mathrm{rsv}(v)$;\quad $X(v) \leftarrow s$;\quad $F \leftarrow F \setminus \{s\}$\;
}
\tcp{Stage 2: front-load large exams}
\ForEach{unassigned $b \in \mathcal{B}$ with $e(b)\ge\tau$, in non-increasing $e(b)$ order}{
    $P \leftarrow F \cap \{s \in \mathcal{S}: s < c\}$ \tcp*{free pre-cutoff slots}
    \If{$P \neq \emptyset$}{
        $s \leftarrow \min P$;\quad $X(b) \leftarrow s$;\quad $F \leftarrow F \setminus \{s\}$\;
    }
}
\tcp{Stage 3: enforce day-load caps}
\ForEach{$d \in \mathrm{dom}(\theta_\mathrm{cap})$}{
    $L \leftarrow \sum_{b':\,X(b')\in\mathcal{S}_d} e(b')$ \tcp*{current day-$d$ enrollment}
    \ForEach{unassigned $b \in \mathcal{B}$, in non-decreasing $e(b)$ order}{
        \If{$F \cap \mathcal{S}_d = \emptyset$ \textnormal{\textbf{ or }} $L + e(b) > \theta_\mathrm{cap}(d)$}{
            \textbf{break}\;
        }
        $s \leftarrow \min(F \cap \mathcal{S}_d)$;\quad $X(b) \leftarrow s$;\quad $F \leftarrow F \setminus \{s\}$;\quad $L \leftarrow L + e(b)$\;
    }
}
\tcp{Stage 4: default placement}
\ForEach{unassigned $b \in \mathcal{B}$}{
    \uIf{$X^0(b) \in F$}{$X(b) \leftarrow X^0(b)$\;}
    \Else{$X(b) \leftarrow \min F$\;}
    $F \leftarrow F \setminus \{X(b)\}$\;
}
\Return $X$\;
\caption{Exam-scheduling heuristic warm start.}
\label{alg:exam-warmstart}
\end{algorithm}

\end{appendices}

\begin{spacing}{1}
\typeout{}
\bibliographystyle{apalike}
\bibliography{References.bib}
\end{spacing}

\end{document}